\documentclass{article}

\PassOptionsToPackage{numbers, compress}{natbib}
 \usepackage[preprint]{neurips_2026}

\usepackage[utf8]{inputenc} 
\usepackage[T1]{fontenc}    
\usepackage{hyperref}       
\usepackage{url}            
\usepackage{booktabs}       
\usepackage{amsfonts}       
\usepackage{nicefrac}       
\usepackage{microtype}      
\usepackage{xcolor}         
\usepackage{graphicx}
\usepackage{subcaption}
\usepackage{wrapfig}
\usepackage{xparse}
\usepackage{xspace}
\usepackage[linesnumbered,ruled,vlined]{algorithm2e}
\usepackage{multirow}
\usepackage{multicol}
\usepackage{adjustbox}
\usepackage{tikz}
\usepackage{comment,url,graphicx,subcaption,relsize}
\usepackage{amssymb,amsfonts,amsmath,amsthm,amscd,dsfont,mathrsfs,mathtools,nicefrac}
\usepackage{float,psfrag,epsfig,color,xcolor,url,hyperref}
\usepackage{epstopdf,bbm,enumitem}

\newtheorem{theorem}{Theorem}
\newtheorem{condition}{Condition}
\newtheorem{lemma}[theorem]{Lemma}

\newif\ifdraft
\drafttrue

\ifdraft

\fi
 \usepackage{makecell}
\renewcommand{\paragraph}[1]{\noindent\textbf{#1}}

\title{Metric Match: A Subset Selection Approach to Evaluating LLM Judge Reliability}

\newcommand*\samethanks[1][\value{footnote}]{\footnotemark[#1]}

\author{%
  Alyssa Unell\thanks{Equal contribution, authors listed alphabetically.} \\
  Department of Computer Science\\
  Stanford University\\
  \texttt{aunell@stanford.edu} \\
  \And
  Natalie Dullerud\samethanks \\
  Department of Computer Science\\
  Stanford University\\
  \texttt{ndulleru@stanford.edu} \\
  \And
  Naomi Boneh \\
  Department of Computer Science\\
  Stanford University\\
  \texttt{naomicyb@stanford.edu} \\
  \And
  Meena Jagadeesan \\
  Department of Computer Science\\
  Stanford University\\
  \texttt{meenaj@seas.upenn.edu} \\
    \And
  Tatsu Hashimoto \\
  Department of Computer Science\\
  Stanford University\\
  \texttt{thashim@stanford.edu} \\
    \And
  Nigam Shah \\
  Department of Medicine\\
  Stanford University\\
  \texttt{nigam@stanford.edu} \\
    \And
  Sanmi Koyejo \\
  Department of Computer Science\\
  Stanford University\\
  \texttt{sanmi@stanford.edu} \\
}

\begin{document}

\maketitle

\setcounter{footnote}{0}

\begin{abstract}
LLM judges are used to reduce the need for costly human labor in evaluating open-ended text generation. However, the reliability of these judges depends critically on their alignment with human raters — a property that itself depends on costly human annotations. 
In this work, we develop a method (\textbf{Metric Match})  for estimating correlation-based reliability metrics of LLM judges from limited annotations. \textbf{Metric Match} selects a subset of samples for human annotation such that the subset matches the population reliability metric with respect to acquired synthetic labels. 
We empirically show that \textbf{Metric Match} achieves a win-rate of 0.838 against random subset selection across four different correlation metrics and 15 datasets, with an 18.7\% decrease in average estimation error and reduces annotation needs by 32.5\%. We provide a cost model and highlight a medical case study where our method saves \$1,041.67 compared to random selection for expert annotation. Further, we shift our task from reliability estimation to reliability classification of whether a given judge is above a deployment threshold, outperforming random selection with \textbf{Metric Match}.
All project code is publicly available\footnote{\url{https://github.com/som-shahlab/MetricMatch}}, and we additionally provide an installable package for ease of use.
\end{abstract}

\section{Introduction}
\label{sec:introduction}

Large language models (LLMs) are increasingly used for text generation tasks, but their rapid adoption has outpaced our ability to evaluate them at scale~\citep{importance, importance1, textgen, scalable}. Accordingly, the LLM judge framework~\citep{llmjudge}, in which one LLM evaluates the outputs of another model, has emerged as a scalable alternative to human annotation. Scalability gains are particularly relevant in expert-oriented contexts such as healthcare~\citep{asgari2025framework}, where human labeling is slow and expensive.  Recent work has explored this direction through human-labeled benchmarks~\citep{dubois2024alpacaeval} and reference-free evaluation methods~\citep{judgebench}.
 
To responsibly deploy LLM judges in high-stakes domains such as healthcare \citep{li2026scopingreviewllmasajudgehealthcare}, it is necessary to evaluate the reliability of LLM-generated annotations with respect to human labels. Specifically, this evaluation serves as a signal to practitioners on whether the LLM judge can reliably replace a costly human annotator. LLM judge reliability \citep{medhelm, judgereview, icc_proof} is often measured using statistical measures in the inter-rater reliability literature~\citep{icc, krippendorff1970estimating} and standard correlation coefficients~\citep{spearman1904proof, kendall1948rank}. However, a key challenge is that these metrics require human annotations in order to calculate the associated reliability score for the target LLM-judge system, creating a bottleneck for evaluation.
In fact, LLM judge evaluation encounters the same scalability problem that the LLM-as-a-judge framework aims to solve in the first place: human annotations on the full datasets are required to exactly compute reliability. 

In order to mitigate this scalability issue, standard approaches aim to estimate judge reliability using a limited budget for human annotations, focusing on producing an unbiased estimator. One approach is to collect annotations on a randomly chosen subset for estimation of the reliability metric, which produces an unbiased estimate~\citep{cochran1977sampling}.
Random selection or classical statistical sampling permits straightforward finite-sample analysis of the metric estimates. 
Another approach is to generate synthetic labels using another LLM judge and then to correct the systemic bias incurred by these labels. The bias is typically estimated on a randomly chosen subset, and modern approaches such as Prediction-Powered Inference (PPI)~\cite{angelopoulos2023ppi} combine bias correction with the construction of a confidence set.  

In this paper, we take a different perspective: rather than aiming to obtain an unbiased estimator for judge reliability, our goal is to predict judge reliability in order to minimize estimation error. In low-annotation regimes, high variance can render an unbiased estimate practically uninformative, making minimization of total estimation error the more relevant objective. We find that for this estimation task, it is useful to move outside of the set of unbiased estimators. Specifically, while standard approaches collect human annotations on a randomly chosen subset, we leverage the structure of the reliability metric on the synthetic labels across items. This structure in turn determines which subset of items to annotate. 

Our main contribution is a new estimation approach (\textbf{Metric Match})  for evaluating LLM judge reliability, which combines limited human annotations with access to synthetic labels from other LLMs. \textbf{Metric Match} leverages a novel subset selection approach: we collect human annotations from a carefully constructed subset which most closely matches the full population in terms of inter-model reliability between the LLM judge scores and the synthetic labels. This approach uses the inter-model metric to guide subset selection for the estimate of the human-model metric of interest. 

We empirically evaluate \textbf{Metric Match} across a diverse set of models (Claude-3.5-Sonnet \cite{claude}, GPT-4.1 \cite{gpt}, GPT-5 \cite{gpt5}, Deepseek-R1 \cite{deepseek}, and Gemini-2.5-pro \cite{gemini}), and datasets (HANNA \cite{hannastories}, MedVAL \cite{medval}, SummEval \cite{summeval}, and MSLR \cite{mslr}). In each context, we consider different sizes of sampling budgets, and different correlation metrics (ICC \cite{icc}, Krippendorff's $\alpha$ \cite{krippendorff1970estimating}, Spearman's $\rho$ rank correlation \cite{spearman1904proof}, and Kendall's $\tau$ rank correlation \cite{kendall1948rank}). 
\textbf{Our results are as follows:}
\begin{enumerate}[leftmargin=*]

    \item Estimation error: 
    We empirically show that \textbf{Metric Match} outperforms baselines such as annotating on randomly collected subsets, bias correction, and stratified sampling (Figure \ref{fig:average_estimation_error}). We decrease estimation error by an average of 18.7\% when compared to random selection. This results in a reduced annotation requirement of 32.5\% (Figure~\ref{fig:budget_equiv_5}). We provide a cost model to calculate the impact of improved estimation ability, showing cost savings of up to \$1,041.67 for a given dataset, MedVAL~\citep{medval}.

    \item Win-rate evaluation: We then perform a systematic comparison of \textbf{Metric Match} against random selection, the de facto approach in practice. We observe an average win rate of 0.838 against random selection, when estimation error is averaged across contexts and budgets. We also find the average consistently exceeds $0.65$ for each distinct budget and metric (Table \ref{subtab:macro_win_rate}). We perform a similar analysis for a more fine-grained win-rate at a per-trial level (Table \ref{subtab:micro_win_rate}). 

    \item Reliability classification: Finally, we turn to the downstream task of reliability classification. Specifically, a practitioner may use an LLM judge only if the estimated reliability coefficient is above a pre-specified deployment threshold. When we shift our task from reliability estimation to reliability classification, \textbf{Metric Match} has a win rate of 0.652 compared to random selection (Table~\ref{tab:threshold_win_rate_metric}). 
    
\end{enumerate}

Accordingly, our work constitutes a substantial step towards scalable LLM judge evaluation, allowing for practitioners to accelerate judge development and improve early failure detection, while aligning with human preferences and maintaining reliability estimation accuracy with fewer annotations. 
\section{Related Works}
\label{sec:related_works}

\subsection{LLM Evaluation and Human Annotation}
Evaluating the outputs of large language models is a longstanding challenge, particularly as these systems are deployed in open-ended and domain-specific settings \citep{importance, importance1}. As model success moves away from traditional multiple-choice evaluation style and into messier, real world environments, evaluation of model success becomes more nuanced than current lexical approaches can capture \cite{bleu, rouge, novikova-etal-2017-need}. Human evaluation has historically served as the gold standard in these instances due to its sensitivity to nuanced qualities such as coherence, factuality, and appropriateness while also  capturing a notion of target quality that we aim to leverage in model alignment \citep{textgen, alignment}. However, human annotation is costly and difficult to scale, especially in domains such as medicine and law, where annotators require expert knowledge \citep{medhelm}. Expert oncologist labeling can be up to \$500 per hour, and with some tasks taking multiple hours with multiple modes of disagreement, it quickly highlights the limitations of human reliance for LLM output evaluation \cite{cancerguide}. Prior work has examined various dimensions of annotation quality, including inter-annotator agreement, annotator bias, and the reliability of crowd-sourced labels, highlighting the problems that arise even if we had unlimited access to human annotations regarding internal human disagreement \citep{judgereview, clark-etal-2021-thats}. Recent work has explored how practitioners can extend these metrics to allow weaker annotators to enhance the scale of stronger annotators \cite{bowman2022measuringprogressscalableoversight, weaktostrong, kenton2024on}. 

\subsection{LLM-as-a-Judge for Scalable Evaluation}
The \emph{LLM-as-a-judge} paradigm, popularized by \citet{llmjudge}, employs LLMs as automated evaluators to score or rank the outputs of other models. While this approach offers high throughput, subsequent research has identified critical failure modes, including position bias, verbosity bias, and self-preference bias \citep{llmjudge, judgereview}. To mitigate these issues, recent benchmarks such as AlpacaEval \citep{dubois2024alpacaeval} introduce length-controlled metrics via regression-based debiasing, while JudgeBench \citep{judgebench} focuses on objective correctness in knowledge-heavy domains where human stylistic preferences may mislead automated judges. Further, comparison of judge responses to a subset of human responses with associated reliability metrics, such as ICC, Krippendorf's $\alpha$, Kendall's $\tau$, and Spearman's $\rho$, serves as signals regarding aforementioned judge biases and shortcomings \cite{metric1,metric2, kendall1948rank, spearman1904proof}. LLM-as-a-judge are frequently compared to human outputs in order to decide whether a system is acceptable to serve as a human proxy or whether further iteration is needed \cite{chiang2023largelanguagemodelsalternative}. This approach leaves open the question of how to use judge outputs effectively when human annotations are scarce.

\subsection{Efficient Sampling and Annotation}

A body of work on bias correction uses a small set of human labels to ``rectify'' the bias of automated predictions. Modern approaches such as Prediction-Powered Inference (PPI)  use a small set of human labels to ``rectify'' the bias of automated predictions, providing valid statistical guarantees on population parameters \citep{angelopoulos2023ppi, feng2026noisy}. Bias correction is typically performed on a random subset of data on which human labels are collected.  The choice of which points to label is itself a variance-reduction problem, and a related line of work frames it through importance sampling: rather than sampling uniformly, points are drawn (or reweighted) according to their expected contribution to the estimator's variance, so that the corrected estimate is more accurate for a fixed labeling budget \citep{chaganty2017importance, zhao2012importance}. 

A handful of works (e.g., \cite{zrnic2024, wu2026efficient}) have explored how to move beyond random selection, and instead construct this subset in an active manner. Such works build a latent factor model on synthetic labels to determine model uncertainty, which then guides active annotation selection. Our approach is similar in motivation, as we use synthetic labels to construct a subset of data points on which to collect human annotations. However, these approaches use model uncertainty estimates to sample from uncertain regions of the distribution as well as learns parameters of latent factor models. These bias correction methods, in addition to PPI and importance sampling, aim to improve the confidence intervals around the estimation, as opposed to optimizing for the point estimation itself.


Our work contributes to a broader body of work on sampling and annotation. To maximize the information of limited annotation budgets, researchers have looked to methodological antecedents in active learning (AL) and optimal experimental design \citep{cochran1977sampling, mackay1992information, Huan2024}. Active learning focuses on iteratively selecting samples with the highest uncertainty or diversity for labeling \citep{sener2018active}, while experimental design approaches optimize the one-step selection of samples to maximize information gain \citep{zrnic2024, wu2026efficient}. 

\section{Preliminaries}
\label{sec:preliminaries}
\begin{figure}
    \centering
    \includegraphics[width=\linewidth]{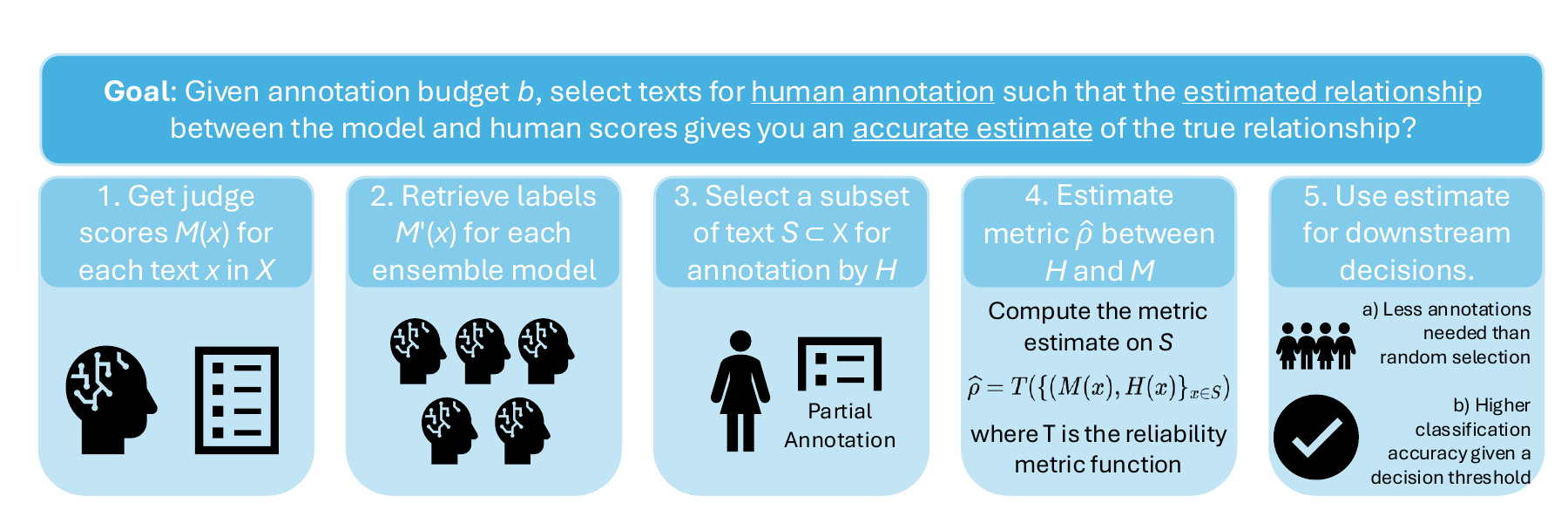}
    \caption{\textbf{Overview of the LLM judge evaluation framework and our approach.} Given text samples $\mathcal{X}$, and judge model $M$ to evaluate, we obtain scores $\{M(x), M'(x)\}_{x\in\mathcal{X}}$ for $M' \in \mathcal{M}$. We select $b$ texts for human annotation. Equipped with a measure of reliability, function $T$, we estimate a target parameter $\hat{\rho}$. We show that our improved estimation of $\hat{\rho}$ leads to downstream impacts.}
    \label{fig:figure_1}
\vspace{-5mm}
\end{figure}

In this section, we outline the problem setup, describe our proposed method (\textbf{Metric Match}) and provide intuition for it, and discuss the experimental setup. 

\subsection{Problem Statement}
\label{subsec:problem_statement}
We study the problem of LLM judge evaluation under limited annotation budgets. Assume we have a large set of text items $\mathcal{X}$, and let $H: \mathcal{X} \rightarrow \mathcal{Y} := [1, K]$, denote the ground-truth human scoring function from items to scores, where $K$ is the greatest possible score. The human scoring function may correspond to a single human rater or an average of multiple raters. Let $M: \mathcal{X}  \rightarrow \mathcal{Y}$ denote the scoring function produced by the LLM judge. To generate synthetic labels, we have access to a family of models $M' \in \mathcal{M}$, each of which yields a synthetic scoring function $M': \mathcal{X} \rightarrow \mathcal{Y}$. Note that the average human rater may produce non-integers, whereas the LLM judge will always produce integer scores.

We wish to measure the reliability of judge $M$ with respect to human annotator $H$ on textual data. We consider a reliability metric $T: (\mathcal{Y} \times \mathcal{Y})^{|\mathcal{X}|} \rightarrow \mathbb{R}$, which operates on pairs of scores on a set of text samples and outputs a number quantifying the reliability between scorers. Given the chosen metric, we want an estimate of the true (population-level) relationship between $M$ and $H$ on $\mathcal{X}$, as captured by $\rho = T(\left\{(M(x), H(x))\right\}_{x \in \mathcal{X}})$.
With a budget of size $b \geq 1$, we can construct any $S \subseteq \mathcal{X}$ of size $|S| = b$ and obtain human annotations on the subset, $\left\{H(x)\right\}_{x \in S}$. 
Using the selected human annotations on $S$ and unlimited query access to annotations from $M' \in \mathcal{M}$, we provide an estimate $\widehat{\rho}_S$ of $\rho$. 

To exploit the availability of synthetic scoring functions $M': \mathcal{X} \rightarrow \mathcal{Y}$, $M' \in \mathcal{M}$, we introduce \textit{inter-model} (IM) reliability quantities as measured by $T$, which inform \textbf{Metric Match}. Denote the inter-model population reliability between $M$ and $M'$ as $\rho^{\text{IM}}$:
\begin{equation*}
    \rho^{\text{IM}} = T\big(\{M(x), M'(x)\}_{x\in\mathcal{X}}\big)
\end{equation*}
Unlike the human-model population metric, the inter-model population metric is directly calculable, as we can cheaply obtain scores $M(x)$, $M'(x)$ for every $x\in\mathcal{X}$. Similarly, given any subset $S\subseteq\mathcal{X}$, let $\widehat{\rho}^{\text{IM}}_S$ signify the finite sample estimate of $\rho^{\text{IM}}$ on $S$, i.e.
\begin{equation*}
    \widehat{\rho}^{\text{IM}}_S = T\big(\{M(x), M'(x)\}_{x\in S}\big)
\end{equation*}

\subsection{Method: Metric Matching}
\label{sec:methods}

We introduce our method, which we term \textbf{Metric Match}, for subset selection in order to reduce estimation error in reliability evaluation of LLM judges with respect to human raters. We focus on careful subset selection in order to minimize finite sample error given a budget $b$, relying on synthetic scorers $M'\in\mathcal{M}$ to inform our subset construction. The key insight in the Metric Match algorithm (Algorithm~\ref{alg:generic_metric_match}) is to construct sets of text examples $S\subseteq\mathcal{X}$ whose inter-model empirical estimate $\widehat{\rho}_S^{\text{IM}}$ closely resembles the inter-model population metric $\rho^{\text{IM}}$. We characterize the intuition for and present details of our algorithm in the rest of this section.

\begin{algorithm}[ht]
\caption{Metric Match}
\label{alg:generic_metric_match}
\DontPrintSemicolon
\KwData{Metric function $T$; budget $b$; population text data $\mathcal{X}$; score functions $M$, $M'$, from LLM judges $M$ and $M'\in\mathcal{M}$, respectively; $C$, number of candidate subsets over which to search}
\KwResult{Subset $S^{*} \subseteq X$, $\lvert S^{*} \rvert = b$; Estimator $\widehat{\rho}_{S^{*}}$}
Initialize $\delta_{\min} \gets \infty$\;
Initialize $S^{*} \gets \emptyset$\;
$\rho^{\text{IM}} \gets \frac{1}{|\mathcal{M}|}\sum_{M'\in\mathcal{M}}T\big(\{M(x), M'(x)\}_{x\in\mathcal{X}}\big)$\;
\For{$j = 1, \dots, C$}{
    Sample $S_j$ s.t. $\vert S_j\vert = b$ from ${X}$ without replacement\;
    $\widehat{\rho}_{S_j}^{\text{IM}} \gets \frac{1}{|\mathcal{M}|}\sum_{M'\in\mathcal{M}}T\big(\{M(x), M'(x)\}_{x\in S_j}\big)$\;
    $\delta_j = \lvert \widehat{\rho}_{S_j}^{\text{IM}} - \rho^{\text{IM}}\rvert$\;
    \If{$\delta_j < \delta_{\min}$}{
        $S^{*} \gets S_j$\;
        $\delta_{\min} \gets \delta_j$
    }
}
Calculate human-model metric estimate $\widehat{\rho}_{S^{*}} = T\big(Y_{S^{*}}^{(M)},Y_{S^{*}}^{(H)}\big)$\;
\Return{$S^{*}$, $\widehat{\rho}_{S^{*}}$}
\end{algorithm}

The intuition is as follows. In an oracle setting, where $H(x)$ is available for all $x\in\mathcal{X}$, the ideal subset $S\subseteq \mathcal{X}$ of size $b$ to construct for estimation is simply the subset that minimizes the estimation error $\vert\widehat{\rho}_S - \rho\vert$. While we do not operate in an oracle setting, we have access to judges $M'\in\mathcal{M}$, and we can compute $M(x), M'(x)$ for every $x\in\mathcal{X}$. This motivates constructing a subset that minimizes the proxy (inter-model) estimation error $S^{*} := \arg\min_{S\subseteq\mathcal{X}, \vert S \vert = b} \lvert \widehat{\rho}^{\text{IM}}_S - \rho^{\text{IM}}\rvert$. The success of this method implicitly assumes that the \textit{relative} inter-model estimation errors are good enough proxies for the relative human-model estimation errors: we formalize this in Section~\ref{subsec:rank_connection}.
Given this subset, we compute the estimator on the human-model metric $\widehat{\rho}_{S^{*}} := T\big(\{M(x), H(x)\}_{x\in S^{*}}\big)$. 


There remain two important details that represent parametric choices about our algorithm: 
\begin{itemize}[leftmargin=*]
    \item \textbf{Aggregation across synthetic labels:} Our selection procedure thus far has assumed there is a single synthetic scorer. However, we have access to a full ensemble of scorers $\mathcal{M}$. We compute an inter-model metric for each pair $(M, M')$ over $M' \in \mathcal{M}$, and average the pairwise computations, resulting in:
\begin{equation*}
    \rho^{\text{IM}} = \frac{1}{|\mathcal{M}|}\sum_{M'\in\mathcal{M}} T\big(\{M(x), M'(x)\}_{x\in\mathcal{X}}\big) \qquad \rho_S^{\text{IM}} = \frac{1}{|\mathcal{M}|}\sum_{M'\in\mathcal{M}} T\big(\{M(x), M'(x)\}_{x\in S}\big)
\end{equation*}
We ablate with an alternate aggregation approach in Appendix~\ref{app:average_then_pairwise}. 
\item \textbf{Candidate Subset Selection:} To solve $S^{*} := \arg\min_{S\subseteq\mathcal{X}, \vert S \vert = b} \lvert \widehat{\rho}^{\text{IM}}_S - \rho^{\text{IM}}\rvert$, we would need to search over ${|\mathcal{X}| \choose b}$ subsets, which is computationally expensive. To reduce the computational overhead, we sample $C$ i.i.d. candidate subsets, $S_1,\dots, S_C \sim \mathcal{X}^b$, from which $S^{*}$ is selected as follows $S^{*} := \arg\min_{S_1,\dots, S_C} \lvert \widehat{\rho}^{\text{IM}}_{S_j} - \rho^{\text{IM}}\rvert$. The number of candidate subsets to consider constitutes a hyperparameter of \textbf{Metric Match}. In our experiments, we set $C = 20$. We vary $C$ and report performance results in Appendix~\ref{app:subsec:n_candidates}. We find that even at relatively small $C$ ($C = 10$), we see gains in performance as measured by win-rate and estimation error. See further analysis of $C$ in Section~\ref{subsec:rank_connection}.
\end{itemize}

\subsection{Experimental Setup}

\paragraph{Baselines.} We compare to the following simple baselines: random selection, stratified sampling, and random selection with a bias correction component.\footnote{Due to the structure of the reliability metrics that we consider, we note that even random sampling and bias correction may not be unbiased (Appendix \ref{app:metric_functions}). However, we take these estimators to be baselines in the context of our goal of minimizing estimation error.}
\begin{itemize}[leftmargin=*]
    \item \textit{Random selection} A classic approach to estimation of $\rho$ is to select $b$ samples i.i.d. from $\mathcal{X}$ to construct $S$, and estimate $\widehat{\rho}_S = T\big(\{M(x), H(x)\}_{x\in S}\big)$. Random selection remains the de facto method in practice. However, random selection incurs high variance when $b$ is small through finite sample error. 
   
    \item \textit{Stratified sampling} Stratified sampling induces a uniform distribution over the judge score space in order to construct $S$. In other words, given $b$, sample $b / K$ text examples from each of $\{x \in \mathcal{X} \mid M(x) = k\}$ for $k = 1,\dots, K$ to ensure full coverage of the score set $\mathcal{Y}$ in the sampled judge scores.\footnote{In practice, we partition $\mathcal{Y}$ into quantiles, and sample uniformly from the quantiles. As LLM judge scores are from a discrete set of integers, the procedure is effectively equivalent. See further implementation details in Appendix~\ref{app:baseline_implementation}.}
     \item \textit{Bias correction (BC)} In bias correction, we leverage the human annotations for the purpose of debiasing the synthetic labels. We use finite sample error calculated on the inter-model reliability metric as a stand-in for finite sample error in the human-model metric.
    Formally, we sample $S\sim\mathcal{X}^b$, and calculate estimate $\widehat{\rho} := \underbrace{\rho^{\text{IM}}}_{\text{Inter-model metric}} + \underbrace{\widehat{\rho}_S - \widehat{\rho}^{\text{IM}}_S}_{\text{Bias correction term}}$.
\end{itemize}

We use $\widehat{\rho}$ to denote a generic estimator for $\rho$ under any method (implicitly with access to $b$ samples). Note that in stratified sampling, the important component of the estimator is the construction of subset $S$, and given $S$, $\widehat{\rho} = \widehat{\rho}_S$. The bias correction estimator has a different structure, due to incorporation of synthetic labels and an additive correction term.

\paragraph{Evaluation procedure.} To evaluate \textbf{Metric Match} (Section~\ref{sec:methods}), we measure the \textbf{estimation error} compared to the baselines and also measure the \textbf{win-rate} of \textbf{Metric Match} compared to the random selection baseline. 
\begin{itemize}[leftmargin=*]
    \item 
\textbf{Estimation error.} Given a method $A$, we directly compute estimation error as $\epsilon^A = \vert \widehat{\rho} - \rho\vert$, where $\widehat{\rho}$ indicates the estimator given by the method to be evaluated. We compute the average estimation error over $N=40$ trials. 
Let $\epsilon^{\text{rand}}$ be the estimation error incurred by random selection. 
\item \textbf{Win-rate.} A ``win'' for any method corresponds to an estimation error smaller than random estimation error. Due to the ubiquitous use of random selection in practice, we focus on the win-rate against random selection. 
We consider two variants of the win-rate.
\begin{itemize}[leftmargin=*]
    \item \textbf{Micro-average win-rate:} Over $N$ trials (with estimation errors $\epsilon_i$, $i=1,\dots, N$), the micro-average win-rate is equal to $\frac{1}{N} \sum_{i=1}^N \mathbbm{1}[\epsilon^{\text{method}}_i < \epsilon^{\text{rand}}_i]$. We compute this statistic for each of the 75 dataset/evaluation axis/judge model contexts. We also report the average of the micro-average win-rate over the 75 contexts as a summary metric. 
    \item \textbf{Macro-average win-rate:} This variant captures the win signal over averaged errors, calculated as $\mathbbm{1}\Big[\frac{1}{N}\sum_{i=1}^N\epsilon^{\text{method}}_i < \frac{1}{N}\sum_{i=1}^N\epsilon^{\text{rand}}_i\Big]\in\{0,1\}$. We compute a win signal for each of the 75 contexts, and average the win signal to produce the macro-average win-rate as a summary metric. 
\end{itemize}
\end{itemize}

\paragraph{Reliability metrics.} To capture notions of reliability for evaluation of an LLM judge, we consider the following reliability metrics: intraclass correlation coefficient (ICC)~\cite{icc}, Krippendorff's $\alpha$~\cite{krippendorff1970estimating}, Spearman's $\rho$ rank correlation~\cite{spearman1904proof}, and Kendall's $\tau$ rank correlation~\cite{kendall1948rank}. Based on previous literature~\cite{medhelm, chiang2023largelanguagemodelsalternative, liu2023g}, these metrics encompass a simple suite of correlative measures between LLM judge scores and human scores. 
We provide formulae for each metric computation and comparative exposition on these metrics in Appendix~\ref{app:metric_functions}.

\paragraph{Datasets and axes.} We provide empirical evaluations on 4 diverse real-world datasets: MSLR \citep{mslr}, HANNA \citep{hannastories},  MedVAL \citep{medval}, and SummEval \citep{summeval}. Each dataset is annotated along multiple axes (e.g., coherence, relevance, surprise) with a total of 15 sets of human annotations by which to evaluate sampling performance, as shown in Table~\ref{tab:datasets}. All datasets contain samples that are annotated by more than one human; we take the average score of a subject between total raters as the true score and compare it to the LLM judge score. All datasets use Likert scaling with a range of 1--5 for SummEval and HANNA, 0--2 for MSLR, and 1--4 for MedVAL \citep{likert}. The full range of datasets, axes, and raters per dataset is presented in Appendix~\ref{app:dataset}. The evaluation axes span from more objective to more subjective scores, capturing a broad spectrum of evaluation types. We truncate each dataset to contain $300$ human-labeled samples and perform evaluations on subsets of this annotation set. We perform the evaluations over $40$ trials. Therefore, we evaluate in a total of $5 \times 15 = 75$ contexts over all datasets, axes and target judge models.

\paragraph{Judges and synthetic labels.} Our model suite consists of Claude-3.5-Sonnet \cite{claude}, GPT-4.1 \cite{gpt}, GPT-5 \cite{gpt5}, Deepseek-R1 \cite{deepseek}, and Gemini-2.5-pro \cite{gemini}, with additional model ablations in Appendix~\ref{subsec:small_models}. We run experiments where each model serves as the reference judge model $M$ and the remaining models comprise the synthetic label collection $\mathcal{M}$. 

\section{Estimation Evaluation}
\label{sec:win_rate}

In this section, we discuss empirical results for metric matching. First, we compare the estimation error of \textbf{Metric Match} against baselines and compute annotation savings of using \textbf{Metric Match} instead of random selection (Section~\ref{subsec:estimation_err}).  Then, we compute a win-rate of \textbf{Metric Match} against random selection and discuss a connection between win-rate and subset estimation error rank (Section~\ref{subsec:win_rate_results}). 
Finally, we evaluate \textbf{Metric Match} on an  alternate task of reliability classification given a deployment threshold, highlighting the practical benefit of improved reliability estimation (Section \ref{subsec:thresh}). 

\begin{figure}[t]
    \centering

    \begin{subfigure}[b]{0.45\textwidth}
        \centering
        \includegraphics[width=\textwidth]{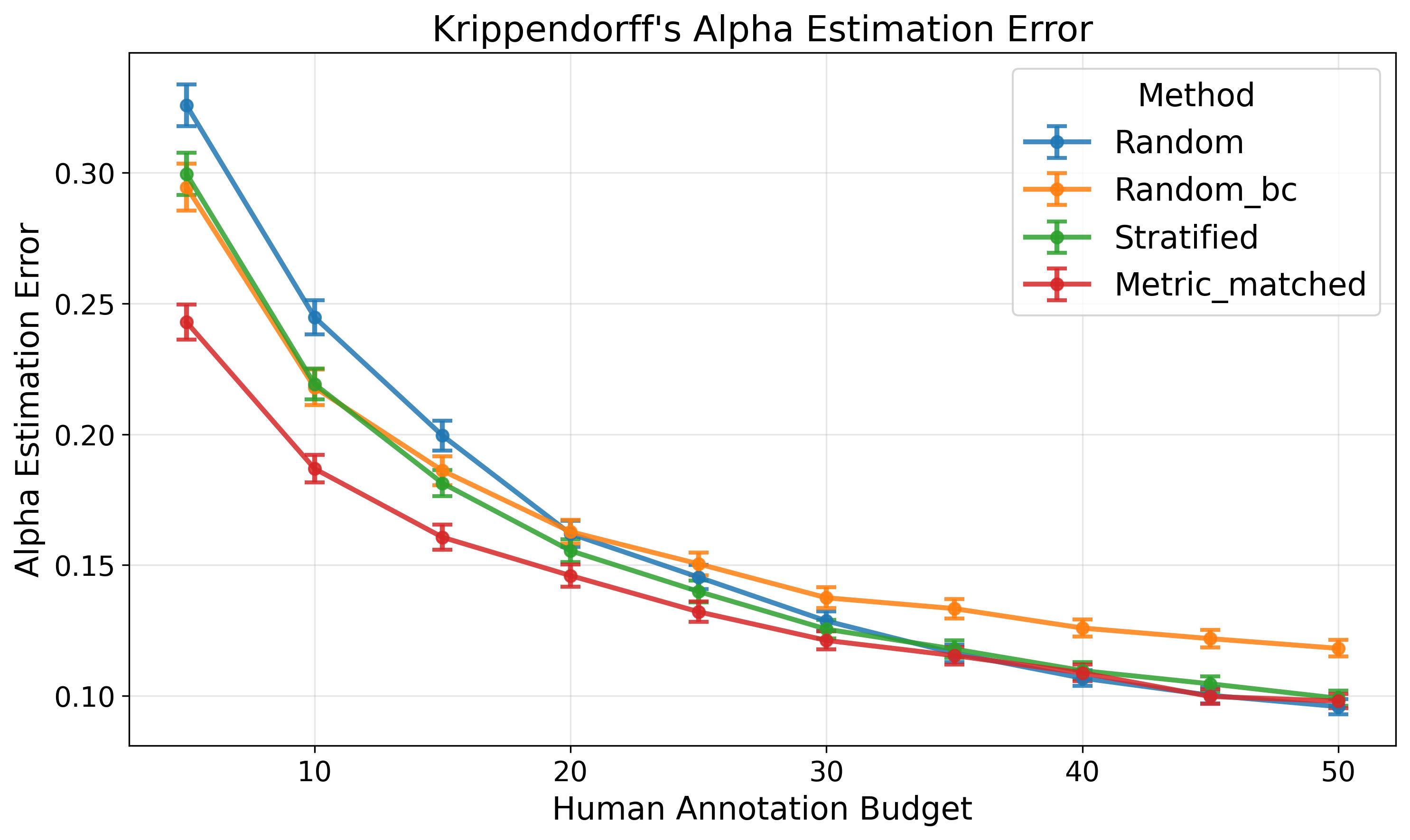}
    \end{subfigure}
    \hfill
    \begin{subfigure}[b]{0.45\textwidth}
        \centering
        \includegraphics[width=\textwidth]{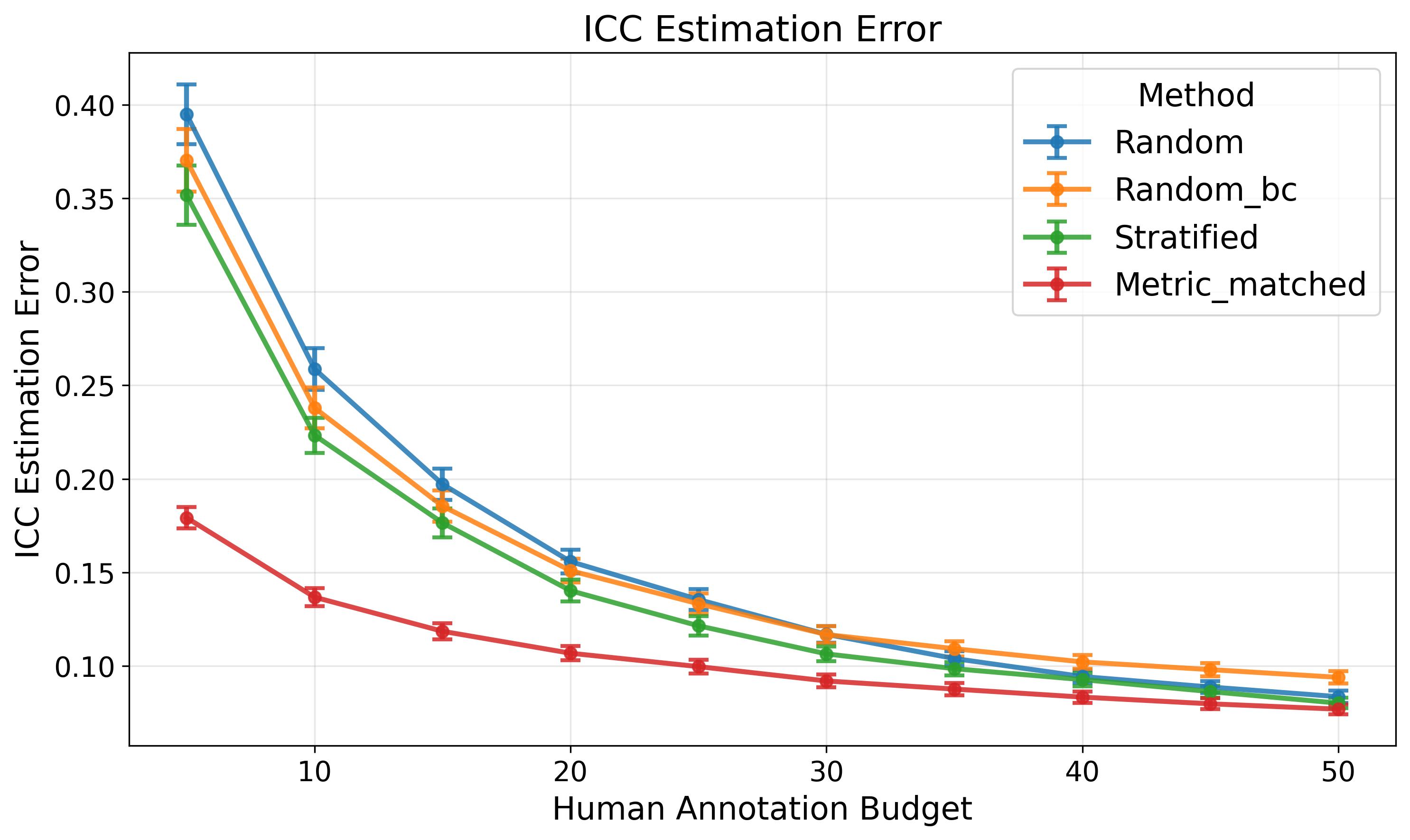}
    \end{subfigure}

    \vspace{0.5cm}

    \begin{subfigure}[b]{0.45\textwidth}
        \centering
        \includegraphics[width=\textwidth]{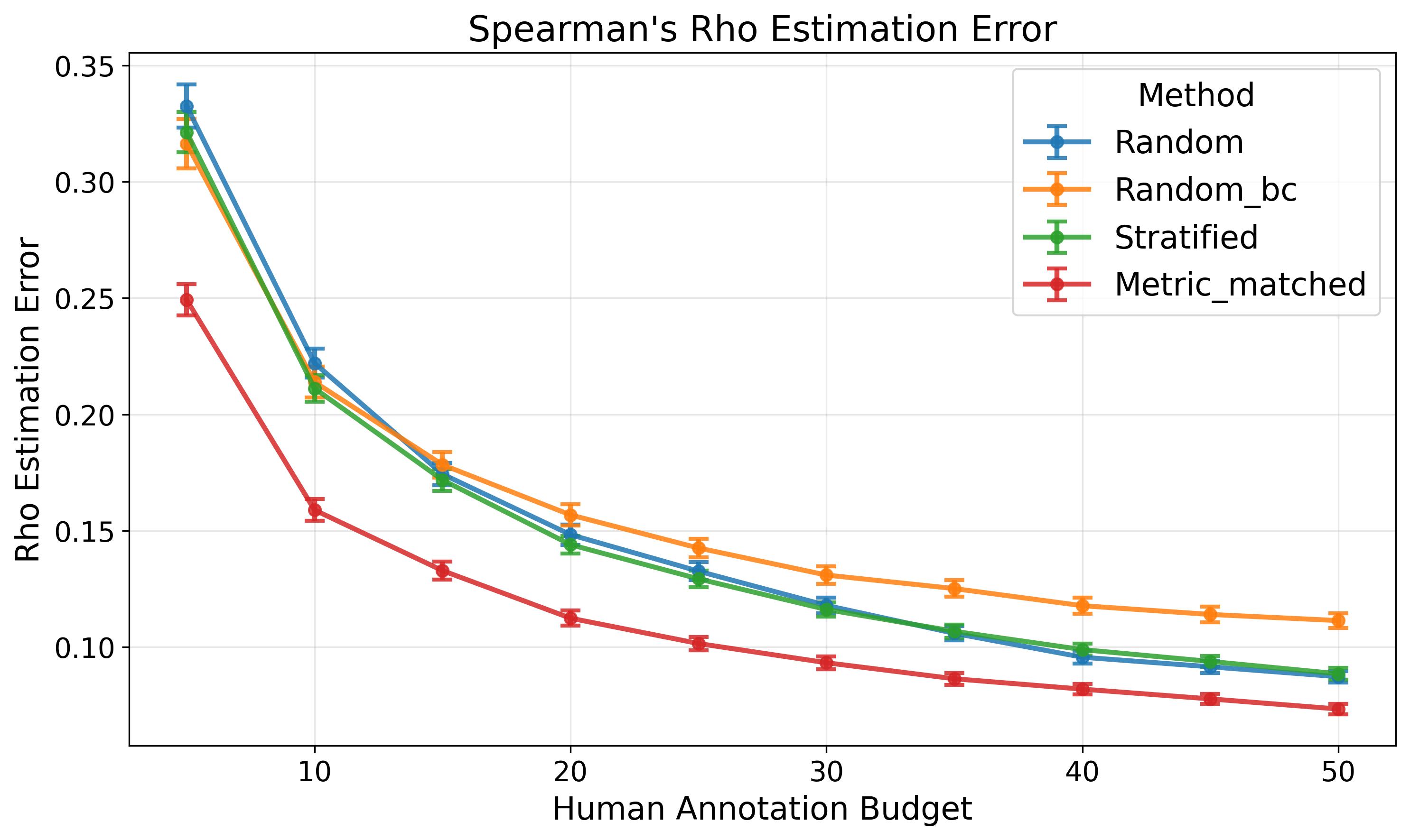}
    \end{subfigure}
    \hfill
    \begin{subfigure}[b]{0.45\textwidth}
        \centering
        \includegraphics[width=\textwidth]{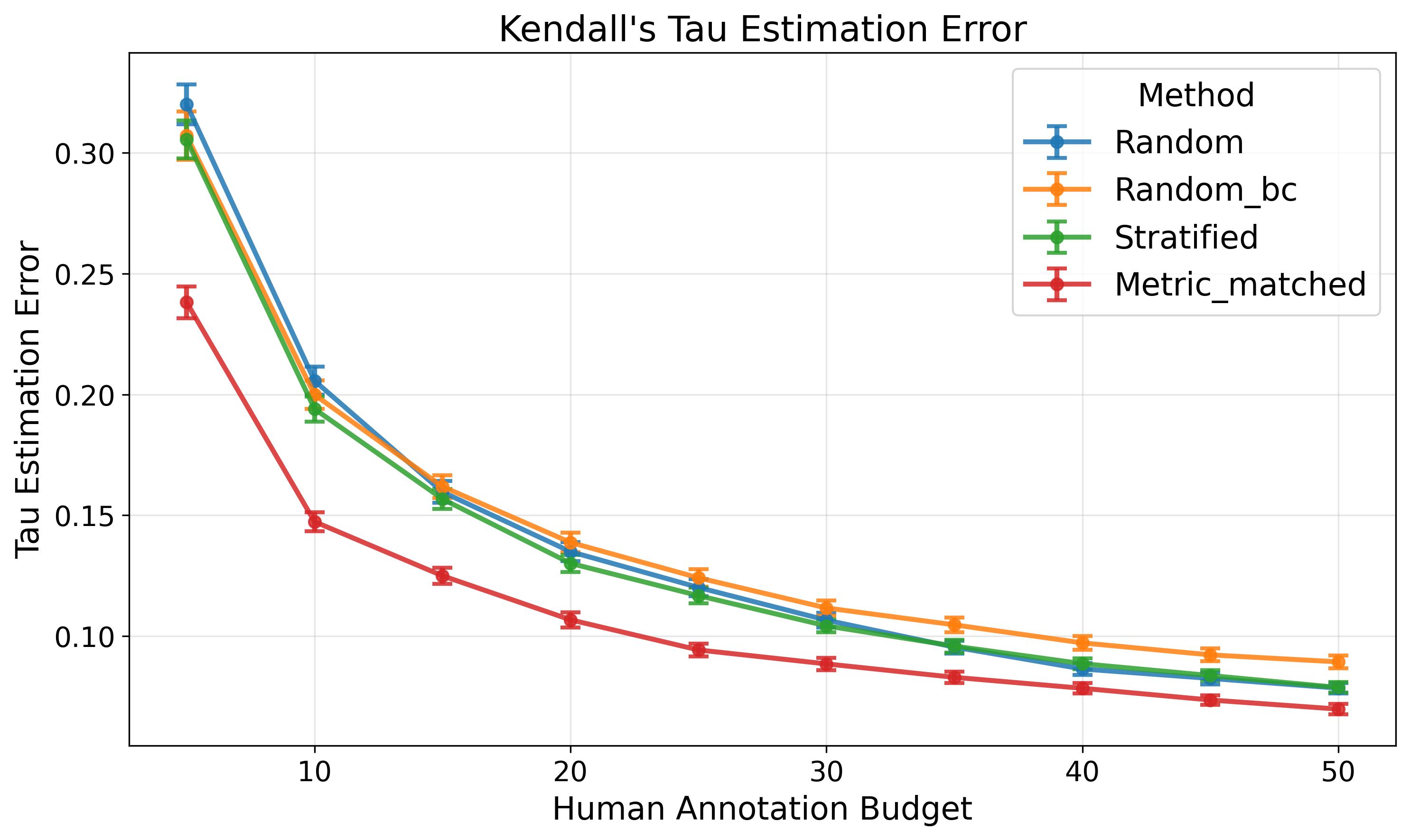}
    \end{subfigure}
        \vspace{0.5cm}

    \caption{\textbf{Metric matched selection outperforms baselines.} We report the estimation error by annotation budget across our suite of target metrics, averaged over all datasets. The relative improved estimation error across all metrics is 18.7\%.}
    \label{fig:average_estimation_error}
\end{figure}

\subsection{Estimation Error}
We quantify the impact of \textbf{Metric Match} on reliability estimation against random selection, showcasing raw improvement in estimation across metrics, datasets, and budgets in Section~\ref{subsubsec:rei}. From this improvement, we calculate relative annotation savings in Section~\ref{subsubsec:ras} and the associated cost savings in Section~\ref{subsubsec:acs}.
\label{subsec:estimation_err}
\subsubsection{Raw Estimation Improvement}
\label{subsubsec:rei}
We highlight in Figure~\ref{fig:average_estimation_error}, the average estimation error of metric matching and the suite of comparable baselines. Across metrics, we see improvement in estimation error with metric matching, most significantly at low budgets, demonstrating the utility of metric matching in labor-constrained domains, where human annotation is extremely costly. Notably, we do not observe a degradation in our performance at higher budgets, which we see in the bias-correction baseline (\textit{Random\_bc}).  We calculate the relative improved estimation error across all metrics as 18.7\%, with disagreggated metrics as follows: 26.7\% for ICC, 9.1\% for Krippendorff's $\alpha$, 21.0\% for Spearman's $\rho$, and 17.9\% for Kendall's $\tau$.

Our results illustrate that for each of the chosen reliability metrics, \textbf{Metric Match} achieves a lower estimation error than baselines which naively collect human annotations. We display disaggregated results for dataset/axis contexts in Appendix~\ref{subsec:not_agg}. In dataset/axis contexts in which metric matching decays performance relative to random selection, we hypothesize that the key assumption on rank correlation between inter-model estimation error and human-model estimation error is violated. We validate this hypothesis empirically in Appendix~\ref{app:rank_estimation_error_connection}, and provide intuitive analysis in Section~\ref{subsec:rank_connection}. 






\begin{figure}[htbp]
    \centering
    \includegraphics[width=.75\linewidth]{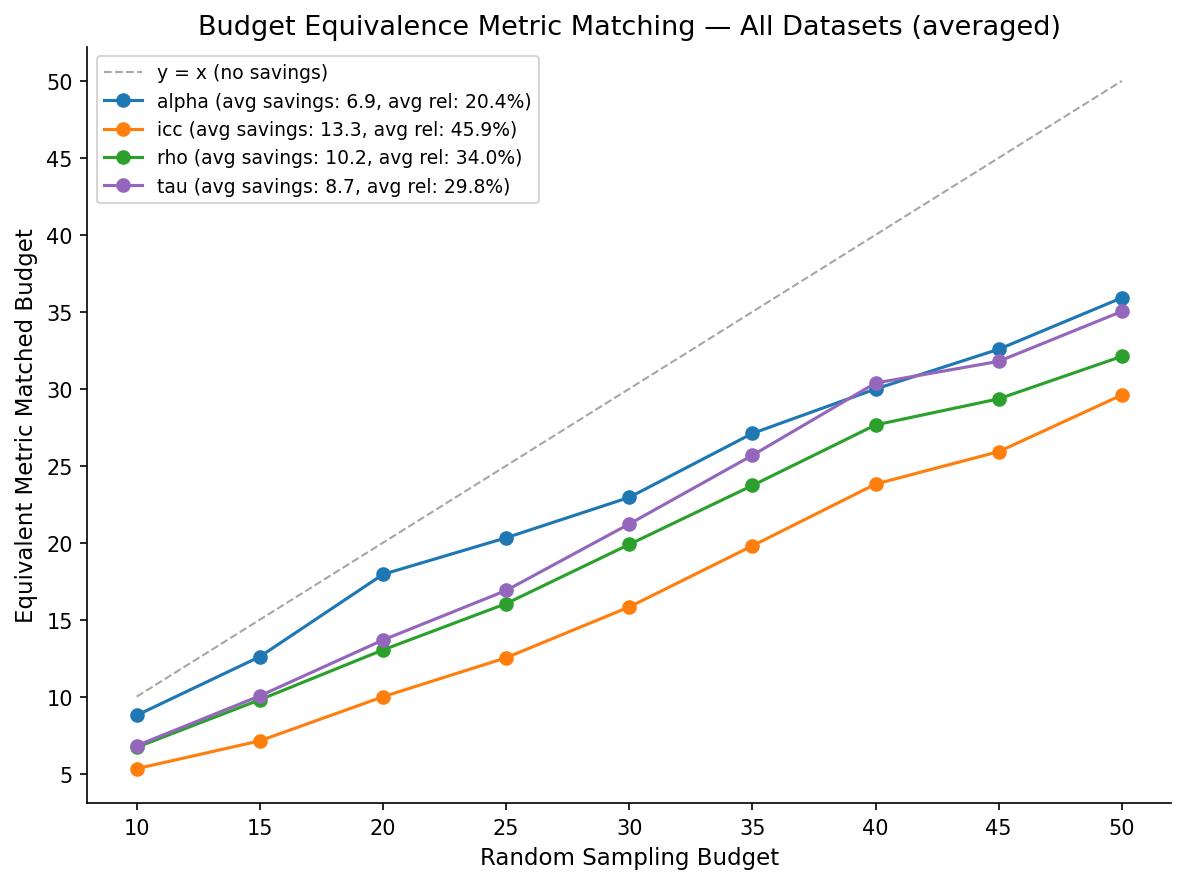}
    \caption{\textbf{Metric matched selection achieves the same estimation error as random with less annotations required.} We highlight the average relative savings over each reliability metric between metric matched selection and random selection, with an average relative improvement of 32.5\%.}
    \label{fig:budget_equiv_5}
\end{figure}

\subsubsection{Relative Annotation Savings}
\label{subsubsec:ras}
We highlight in Figure~\ref{fig:budget_equiv_5} the average annotations saved from using metric matching for sample selection. For each budget given to random selection, we compute the required budget needed for \textbf{Metric Match} to match the estimation error of random selection according to Figure \ref{fig:average_estimation_error}. We plot the relationship between the random selection budget and our required budget. 
As such, a line with a slope of 1 would indicate that \textbf{Metric Match} performs as well as random, while points below that line indicate \textbf{Metric Match} outperforming random selection. We see gains across all metrics observed, with an average relative improvement of 32.5\%. Disaggregated results are reported in Appendix~\ref{subsec:not_agg}.

\subsubsection{Annotation Cost Savings}
\label{subsubsec:acs}
Let $R$ be the random sampling budget, $t$ be the estimated time per annotation (in hours), and $c$ be the cost per unit time of annotation (in dollars per hour). Given $R$, we determine the required annotation budget $M$ by identifying the point at which the estimation error of \textbf{Metric Match} selection equals the annotation error at $R$. We quantify cost savings as:

\begin{equation}
    C(R) = (R - E_m(R)) \cdot t \cdot c
\end{equation}

where $E_m(R)$ denotes the Metric Matched annotation budget that achieves equivalent estimation accuracy to random sampling at budget $R$. For MedVAL, assuming 5 minutes per annotation at a cost of \$500 per annotator hour (as described in \cite{cancerguide}), this becomes for ICC:

\begin{equation}
    \$1{,}041.67 = (50 - 25) \cdot \frac{1}{12} \cdot 500
\end{equation}







\subsection{Win-rate against random selection baseline}
\label{subsec:win_rate_results}
We show in Table~\ref{tab:est_win_rate} the win-rate over trials between \textbf{Metric Match}'s estimation error with the target parameter and random's estimation error. 
In terms of macro win-rate, we beat random with an average win rate across metrics of 0.838, indicating robustness of metric matching for estimation error across metrics. In micro win-rate, we beat random in 53.8\% of empirical trials. 
We include disaggregated win rate tables for individual axis performance in Appendix~\ref{subsec:not_agg}, as well as win rates against the other two baselines in Appendix~\ref{subsec:strat}. 

\begin{table}[t]
    \centering
    \caption{\textbf{Metric matched selection beats random selection for minimizing estimation error.} Comparison of Macro (left) and Micro (right) win rates across different budgets. Macro win rates achieve an average of 0.838, while micro win rates achieve an average of 0.538.}
    \label{tab:est_win_rate}
    \begin{subtable}{0.48\textwidth}
        \centering
        \subcaption{Macro Win Rate}
        \label{subtab:macro_win_rate}
        \begin{tabular}{c|cccc}
            \toprule
            Budget & $\alpha$ & ICC & $\rho$ & $\tau$ \\
            \midrule
            5 & 0.853 & 1.000 & 0.947 & 0.987 \\
            10 & 0.827 & 0.987 & 0.947 & 0.973 \\
            15 & 0.747 & 0.947 & 0.907 & 0.907 \\
            20 & 0.733 & 0.920 & 0.960 & 0.920 \\
            25 & 0.720 & 0.907 & 0.880 & 0.907 \\
            30 & 0.733 & 0.893 & 0.867 & 0.893 \\
            35 & 0.653 & 0.853 & 0.893 & 0.773 \\
            40 & 0.653 & 0.840 & 0.800 & 0.760 \\
            45 & 0.667 & 0.800 & 0.800 & 0.747 \\
            50 & 0.653 & 0.773 & 0.787 & 0.693 \\
            \midrule
            Avg & 0.724 & 0.892 & 0.879 & 0.856 \\
            \bottomrule
        \end{tabular}
    \end{subtable}
    \hfill 
    \begin{subtable}{0.48\textwidth}
        \centering
        \subcaption{Micro Win Rate}
        \label{subtab:micro_win_rate}
        \begin{tabular}{c|cccc}
            \toprule
            Budget & $\alpha$ & ICC & $\rho$ & $\tau$ \\
            \midrule
            5 & 0.574 & 0.641 & 0.580 & 0.575 \\
            10 & 0.569 & 0.616 & 0.595 & 0.584 \\
            15 & 0.537 & 0.591 & 0.562 & 0.552 \\
            20 & 0.507 & 0.559 & 0.556 & 0.546 \\
            25 & 0.511 & 0.554 & 0.560 & 0.555 \\
            30 & 0.501 & 0.548 & 0.551 & 0.535 \\
            35 & 0.476 & 0.522 & 0.540 & 0.521 \\
            40 & 0.467 & 0.511 & 0.529 & 0.505 \\
            45 & 0.476 & 0.505 & 0.527 & 0.509 \\
            50 & 0.465 & 0.484 & 0.528 & 0.509 \\
            \midrule
            Avg & 0.508 & 0.553 & 0.553 & 0.539 \\
            \bottomrule
        \end{tabular}
    \end{subtable}
\end{table}


Importantly, while macro win-rate effectively constitutes a measure of the frequency (across contexts) with which our \textit{expected} estimation error beats \textit{expected} random error, micro win-rate is an empirical estimate of the probability of dominating random selection in a single trial. We see that in both cases, across budgets, and metrics, our win rate is greater than $0.5$.

Improved performance with \textbf{Metric Match} generalizes across model scales (see Appendix~\ref{subsec:small_models} for small model results) and outperforms random sampling on alternative metrics including mean squared error (Appendix~\ref{subsec:mse}). We additionally show results when matching on complementary metrics in Appendix~\ref{subsec:var_matching}, highlighting that alternative signals of inter-model reliability may be helpful for optimal subset selection, as well.

\subsubsection{Analysis of win-rate and expected subset rank}\label{subsec:rank_connection}

As noted in Section~\ref{sec:methods}, the performance of metric matching relies on the joint distribution over the ranks of the inter-model and human-model estimation errors in subsets $S \subseteq \mathcal{X}$. Here, we clarify the precise relationship between micro-average win-rate and the expected rank of subset $S^{*}$ over possible subsets with respect to the human-model estimation error. Consider the candidate subsets $S_1,\dots, S_C \sim\mathcal{X}$: let $\epsilon_j := |\widehat{\rho}_{S_{j}} - \rho|$ and $\epsilon_j^{\text{IM}} := |\widehat{\rho}^{\text{IM}}_{S_{j}} - \rho^{\text{IM}}|$, and $j^{*}$ be the index of $S^{*}$ in $\{1,\dots, C\}$. The expected micro-average win-rate of metric matching is exactly $\mathbb{E}\Big[ \mathbbm{1}\big[\epsilon_{j^{*}} < \epsilon^{\text{rand}}\big] \Big] = \Pr(\epsilon_{j^{*}} < \epsilon^{\text{rand}})$. Let $R$ denote the rank operator with respect to human-model estimation error. Recall that the rank of $S^{*}$ with respect to inter-model estimation error is $1$ by definition. The expected micro-average win-rate of metric matching depends linearly on the expected rank of the $j^{*}$th subset in the human-model, $\mathbb{E}[R[\epsilon_{j^{*}}]]$.
We provide an explicit formula, derivation and empirical validation of this relationship in Appendix~\ref{app:rank_estimation_error_connection}.


\subsection{Reliability Classification}

\label{subsec:thresh}
In practice, a key motivation for estimating judge reliability is to determine whether the judge is sufficient to be used for the downstream task. This corresponds to a classification task of determining whether judge reliability is above a certain threshold $t$ \citep{feng2026noisy}. We evaluate the performance of \textbf{Metric Match} relative to random selection for the purposes of this classification task in Table~\ref{tab:threshold_win_rate_metric}. Given an estimate $\rho$, we derive a prediction based on $1[\rho > t]$. 
We compute a macro-averaged win-rate and micro-averaged win-rate for \textbf{Metric Match} relative to random selection at thresholds of $t \in \{0.6, 0.7, 0.8\}$. Here, a win is when \textbf{Metric Match} produces a correct classification while random selection produces an incorrect classification. We average win rates across thresholds.
\begin{table}[htbp]
    \centering
    \caption{\textbf{Metric matched selection beats random selection for reliability classification task.} Comparison of Macro (left) and Micro (right) win rates across different budgets. Macro win rates achieve an average of 0.652, while micro win rates achieve an average of 0.570.}
    \label{tab:threshold_win_rate_metric}
    \begin{subtable}{0.48\textwidth}
        \centering
        \subcaption{Macro Win Rate}
        \label{subtab:macro_win_rate_threshold}
        \begin{tabular}{c|cccc}
            \toprule
            Budget & $\alpha$ & ICC & $\rho$ & $\tau$ \\
            \midrule
            5 & 0.658 & 0.519 & 0.825 & 0.832 \\
            10 & 0.763 & 0.566 & 0.794 & 0.907 \\
            15 & 0.767 & 0.558 & 0.731 & 0.788 \\
            20 & 0.535 & 0.534 & 0.644 & 0.451 \\
            25 & 0.612 & 0.595 & 0.805 & 0.742 \\
            30 & 0.659 & 0.547 & 0.733 & 0.678 \\
            35 & 0.606 & 0.556 & 0.789 & 0.684 \\
            40 & 0.527 & 0.542 & 0.702 & 0.647 \\
            45 & 0.458 & 0.535 & 0.690 & 0.786 \\
            50 & 0.545 & 0.563 & 0.468 & 0.739 \\
            \midrule
            Avg & 0.613 & 0.551 & 0.718 & 0.725 \\
            \bottomrule
        \end{tabular}
    \end{subtable}
    \hfill 
    \begin{subtable}{0.48\textwidth}
        \centering
        \subcaption{Micro Win Rate}
        \label{subtab:micro_win_rate_threshold}
        \begin{tabular}{c|cccc}
            \toprule
        Budget & $\alpha$ & ICC & $\rho$ & $\tau$ \\
        \midrule
        5 & 0.585 & 0.507 & 0.648 & 0.700 \\
        10 & 0.603 & 0.526 & 0.654 & 0.724 \\
        15 & 0.598 & 0.521 & 0.599 & 0.618 \\
        20 & 0.496 & 0.512 & 0.579 & 0.504 \\
        25 & 0.542 & 0.547 & 0.646 & 0.582 \\
        30 & 0.581 & 0.517 & 0.574 & 0.608 \\
        35 & 0.579 & 0.502 & 0.586 & 0.586 \\
        40 & 0.473 & 0.473 & 0.602 & 0.617 \\
        45 & 0.468 & 0.483 & 0.528 & 0.744 \\
        50 & 0.484 & 0.487 & 0.503 & 0.682 \\
        \midrule
        Avg & 0.541 & 0.508 & 0.592 & 0.637 \\
        \bottomrule
        \end{tabular}
    \end{subtable}
\end{table}

\section{Discussion}
\label{sec:discussion}

We develop a new method, \textbf{Metric Match}, for estimating the reliability of LLM judges using synthetic labels to best select a representative subset for human annotation. Empirically, \textbf{Metric Match} reduces estimation error by an average of 18.7\% relative to random selection, outperforming baselines including bias correction and stratified sampling. This improved estimation ability reduces the number of annotations required to reach a target error by an average of 32.5\%. In a case study on the MedVAL dataset, our cost model translates this reduction into savings of up to \$1,041.67. To characterize the consistency of these gains, we conduct a systematic win-rate comparison against random selection, the de facto approach in practice, observing an average win rate of 0.838 on estimation error when averaged across contexts and budgets. We provide theoretical justification for this estimation approach, proving that under rank alignment assumptions between human-model and inter-model metrics on subsets, metric matching provably outperforms random selection. Finally, we consider an alternative task to reliability estimation as reliability classification, which is defined as a practitioner using the estimated parameter against a deployment threshold to decide whether a judge is fit for use. For this task, \textbf{Metric Match} attains a win rate of 0.652 on the binary accept-versus-reject decision. Together, these results demonstrate that improved estimation of the population-level target parameter enables practitioners to make more accurate decisions using fewer human annotations.

While this work highlights the performance of metric matching for reliability estimation of correlation-based metrics, future work may engage with a wider set of metrics. We include preliminary analysis for Mean Squared Error as a target parameter in Appendix~\ref{subsec:mse}, showing that metric-matching outperforms random selection on non-correlation based metrics, although with smaller annotation gains and absolute estimation error savings. Quantifying improvement a priori remains an open question. Reliability is use-case dependent, and this estimation approach should not be conflated with accuracy. Additionally, we do not address how to improve the judge score nor how to select the optimal judge from the ensemble. These features are assumed static in our set-up, but are more fluid in real world implementation. Finally, this work doesn't address how to incorporate human feedback online to improve estimation actively. Integrating active learning approaches is likely to improve performance beyond our current batch setting, though this constitutes a different problem formulation.

More broadly, \textbf{Metric Match} offers a promising approach for practitioners to validate LLM judges more efficiently, reducing both the cost and time required for expert annotation. We hope that our work serves as a starting point for organizations to iterate faster on judge development, identify failure modes earlier, and to detect misalignment with human preferences while achieving equivalent estimation accuracy with fewer labeled examples. 

\section{Acknowledgments}

 AU acknowledges support from the Stanford Graduate Fellowship and NSF GRFP. MJ acknowledges partial support from a Stanford AI Lab Postdoctoral fellowship. SK acknowledges support by NSF 2046795 and 2205329, IES R305C240046, ARPA-H, the MacArthur Foundation, Schmidt Sciences, and HAI. TH was supported by a grant by HAI, DSO labs, gifts from Open Philanthropy, Amazon, Schmidt Sciences, the Tianqiao and Chrissy Chen Foundation and a grant under the NSF CAREER IIS-2338866, ONR N00014-24-1-2609, and DARPA Cooperative Agreement HR00112520013. This work does not necessarily reflect the position or policy of the government and no official endorsement should be inferred.

\bibliographystyle{plainnat}
\bibliography{references}

\newpage

\appendix
\section{Metric Functions}
\label{app:metric_functions}
Here, we show equations and population information about our chosen metrics. 

\paragraph{Intraclass Correlation Coefficient}
Intraclass Correlation Coefficient (ICC) was originally proposed as an extension to \emph{interclass} correlation coefficient (Pearson's correlation coefficient (PCC)), and measures the extent to which the total variance in observed data is due to differences between groups, rather than within groups. In this perspective, the ICC is understood within the analysis of variance (ANOVA) framework. As opposed to PCC, the data are pooled in the mean calculation. 

In the generic version, ICC is considered a measure that quantifies inter-rater reliability between $k$ raters on $n$ subjects, first introduced as an application of the metric in~\citet{icc}. ICC measures reliability by decomposing the total variance in human evaluations into between-subjects variance and within-subjects error variance. The ICC determines the reliability of ratings by comparing the variability of different ratings of the same individuals to the total variation across all ratings and all individuals. As we only consider two raters, the human and the LLM, we consider the case $k = 2$. 

Analogously, modern ICC estimators derive ICC through the random effects model framework. In the random effects model, $X_{ij}$, rating $j$ on subject $i$, $i\in[n], j\in[k]$, is modeled as
\begin{equation*}
    X_{ij} = \mu + \alpha_i + c_j + \varepsilon_{ij}
\end{equation*}
such that $\mu$ is an unobserved overall mean, $\alpha_i$ is an unobserved random effect shared by all ratings on subject $i$, $c_j$ is an unobserved random effect shared by all subject ratings by rater $j$, and $\varepsilon_{ij}$ is an unobserved noise term. Each class of terms is assumed to be respectively identically distributed with expected value $0$, and the terms are assumed to be uncorrelated. For certain random effects models, either $\alpha_i$ or $c_j$ is neglected or considered fixed. We refer to~\citet{liljequist2019icc} for a comprehensive overview of ICC definitions and derivations relating classical estimators to the random effects model.

In our specific use case, we use a two-way consistency average, i.e. $\mathrm{ICC}(3, k)$ as this formulation treats \emph{raters} as fixed effects, (i.e. $c_j$ is fixed), meaning the same evaluation panel assesses all LLM outputs, and estimates reliability for the average rating across $k$ evaluators rather than individual rater consistency. The numerator ($MS_R - MS_E$) captures the true variance between different LLM responses after removing measurement error, while the denominator represents the total variance in averaged ratings, making $\mathrm{ICC}(3,k)$ particularly sensitive to systematic differences in how evaluators rate different model outputs while accounting for random measurement error within the evaluation process. With random effects model for $\mathrm{ICC}(3, k)$, the population ICC
\begin{equation*}
    \rho = \frac{\sigma^2_{\alpha}}{\sigma^2_{\alpha} + \sigma^2_{\varepsilon}/k}
\end{equation*}
We utilize the associated formula as the ICC metric for our experiments due to the appropriateness of the setting, random effects model, and use in previous empirical work \citep{medhelm, icc_proof, judgereview}. See table below for reproduced formulas from~\citep{liljequist2019icc} for intraclass correlation coefficient. We use ICC$(3, k)$ in all over our analyses. 

\begin{table}[!h]
\tiny
\begin{tabular}{@{}llllp{4.5cm}@{}}
\toprule
\textbf{Name} & \textbf{Notation} & \textbf{Rater Model} & \textbf{Use Case} & \textbf{Formula} \\
\midrule
One-way single & ICC(1,1) & Random & Agreement of 1 random rater & 
$\displaystyle \frac{\text{MS}_{\text{R}} - \text{MS}_{\text{E}}}{\text{MS}_{\text{R}} + (k-1)\text{MS}_{\text{E}}}$ \\

One-way average & ICC(1,k) & Random & Agreement of average random raters & 
$\displaystyle \frac{\text{MS}_{\text{R}} - \text{MS}_{\text{E}}}{\text{MS}_{\text{R}}}$ \\

Two-way absolute single & ICC(2,1) & Random & Absolute agreement of 1 random rater & 
$\displaystyle \frac{\text{MS}_{\text{R}} - \text{MS}_{\text{E}}}{\text{MS}_{\text{R}} + (k-1)\text{MS}_{\text{E}} + \frac{k}{n}(\text{MS}_{\text{C}} - \text{MS}_{\text{E}})}$ \\

Two-way absolute average & ICC(2,k) & Random & Absolute agreement of average raters & 
$\displaystyle \frac{\text{MS}_{\text{R}} - \text{MS}_{\text{E}}}{\text{MS}_{\text{R}} + \frac{1}{n}(\text{MS}_{\text{C}} - \text{MS}_{\text{E}})}$ \\

Two-way consistency single & ICC(3,1) & Fixed & Consistency of 1 fixed rater & 
$\displaystyle \frac{\text{MS}_{\text{R}} - \text{MS}_{\text{E}}}{\text{MS}_{\text{R}} + (k-1)\text{MS}_{\text{E}}}$ \\

Two-way consistency average & ICC(3,k) & Fixed & Consistency of average fixed raters & 
$\displaystyle \frac{\text{MS}_{\text{R}} - \text{MS}_{\text{E}}}{\text{MS}_{\text{R}}}$ \\

Pearson correlation & $r$ & N/A & Correlation only (not agreement) & 
$\displaystyle r = \frac{\sum (x_i - \bar{x})(y_i - \bar{y})}{\sqrt{\sum (x_i - \bar{x})^2 \sum (y_i - \bar{y})^2}}$ \\
\bottomrule
\end{tabular}
\end{table}

\vspace{1em}
\noindent
\textbf{Notation:}
\begin{itemize}
  \item $\text{MS}_\text{R}$: Mean square between targets (rows)
  \item $\text{MS}_\text{C}$: Mean square between raters (columns)
  \item $\text{MS}_\text{E}$: Residual mean square (error)
  \item $n$: Number of targets
  \item $k$: Number of raters
\end{itemize}

\textbf{Formulas:}
\begin{align*}
    \mathrm{MS_R} &= \frac{k}{n - 1} \sum_{i = 1}^n (S_i - \overline{X}_{\mathrm{tot}})^2 \\
    \mathrm{MS_C} &= \frac{n}{k - 1} \sum_{j = 1}^k (M_j - \overline{X}_{\mathrm{tot}})^2 \\
    \mathrm{MS_E} &= \frac{\sum_{i=1}^n \sum_{j = 1}^k (x_{ij} - M_j)^2 - k\sum_{i = 1}^n (S_i - \overline{X}_{\mathrm{tot}})^2}{(n - 1)(k - 1)} \\
    S_i &= \frac{1}{k}\sum_{j = 1}^k x_{ij} \\
    M_j &= \frac{1}{n}\sum_{i = 1}^n x_{ij} \\
    \overline{X}_{\mathrm{tot}} &= \frac{1}{k \cdot n} \sum_{i = 1}^n \sum_{j = 1}^k x_{ij}
\end{align*}

\paragraph{Krippendorff's $\alpha$} Krippendorff's alpha, introduced in \citet{krippendorff1970estimating} is a statistical measure for reliability common in inter-rater reliability literature. The metric operates on the same reliability matrix as intraclass correlation coefficient, $X \in \mathbb{R}^{n \times k}$, such that $X_ij$ denotes the rating from rater $j$ on subject $i$. Unlike ICC, Krippendorff's alpha is not generally positioned in the context of the random effects model, nor the analysis of variance (ANOVA) framework. The measure is highly versatile, designed to handle a variety of data types (nominal, ordinal, interval, and ratio) and common research obstacles including missing values. 

The general form of the coefficient is based on the ratio of observed disagreement amongst raters to the disagreement expected by chance,
\begin{equation*}
\alpha = 1 - \frac{D_o}{D_e}
\end{equation*}
where $D_o$ is the observed disagreement and $D_e$ is the disagreement expected by chance. To define $D_o$ and $D_e$, define a unit $u$ as the ratings of all raters for item $i$:
\begin{itemize}
    \item $n$ be the total number of values assigned to all units.
    \item $m_i$ be the number of coders who responded to subject $i$.
    \item $v, k$ represent specific values (categories) in the data.
\end{itemize}

Then, the observed disagreement, $D_o$ is the average of the distances between all pairs of values assigned to the same units. It is defined as:
\begin{equation*}
D_o = \frac{1}{n} \sum_{i} \frac{1}{m_i - 1} \sum_{v} \sum_{k} n_{iv} n_{ik} \delta^2_{vk}
\end{equation*}
where
\begin{itemize}
    \item $n_{iv}$ and $n_{ik}$ are the number of times values $v$ and $k$ were assigned to unit $i$.
    \item $\delta^2_{vk}$ is a distance function (metric) that depends on the level of measurement (Nominal, Ordinal, Interval, or Ratio).
\end{itemize}
The expected disagreement, $D_e$, is the disagreement that would occur if the values were assigned to units completely at random, constrained only by the overall frequency of each value in the dataset:
\begin{equation*}
D_e = \frac{1}{n(n-1)} \sum_{v} \sum_{k} n_v n_k \delta^2_{vk}
\end{equation*}
where
\begin{itemize}
    \item $n_v$ and $n_k$ are the total number of times values $v$ and $k$ were used across all units.
\end{itemize}

Finally, we turn to the distance functions ($\delta^2_{vk}$). The power of $\alpha$ lies in the metric $\delta^2_{vk}$, which accounts for the magnitude of disagreement:
\begin{itemize}
    \item \textbf{Nominal:} $\delta^2_{vk} = \begin{cases} 0 & \text{if } v=k \\ 1 & \text{if } v \neq k \end{cases}$
    \item \textbf{Ordinal:} $\delta^2_{vk} = \left( \sum_{g=v}^k n_g - \frac{n_v + n_k}{2} \right)^2$
    \item \textbf{Interval:} $\delta^2_{vk} = (v - k)^2$
    \item \textbf{Ratio:} $\delta^2_{vk} = \left( \frac{v - k}{v + k} \right)^2$
\end{itemize}

Substituting $D_o$ and $D_e$ back into the original formula:
\begin{equation*}
\alpha = 1 - (n-1) \frac{\sum_i \frac{1}{m_i - 1} \sum_v \sum_k n_{iv} n_{ik} \delta^2_{vk}}{\sum_v \sum_k n_v n_k \delta^2_{vk}}
\end{equation*}
This formula ensures that when agreement is perfect, $D_o = 0$ and $\alpha = 1$. When disagreement matches chance, $D_o = D_e$ and $\alpha = 0$.

\paragraph{Spearman $r$} Spearman's correlation coefficient, which we denote as $r$, evaluates the monotonic relationship between two rank variables. In particular, let $X$ and $Y$ be random variables with paired samples $(X_i, Y_i)$ and rank variables $R[X_i], R[Y_i]$, $i = 1,\dots, n$. Spearman's correlation coefficient is the Pearson correlation coefficient on the \textit{rank} variables. Unlike Pearson correlation, which measures linear relationships, Spearman correlation measures the strength and direction of the association between two ranked variables. The equation representing Spearman's rank correlation is shown below:
\begin{equation*}
    r = \frac{\text{Cov}[R[X], R[Y]]}{\sigma_{R[X]}\sigma_{R[Y]}}
\end{equation*}
If there are no ties in ranks, the coefficient can be calculated as
\begin{equation*}
    \rho = 1 - \frac{6 \sum d_i^2}{n(n^2 - 1)}
\end{equation*}
where $d_i = R[X_i] - R[Y_i]$, $i = 1,\dots, n$.

\paragraph{Kendall's $\tau$} Kendall's tau rank correlation measures the ordinal association between two variables based on the similarity of the orderings of the data. The measure is a rank correlation measures as in Spearman's $r$, and operates on ranks of pairs of random variables, $(X_i, Y_i)$, $i = 1,\dots, n$. Differing from Spearman's $r$, Kendall's $\tau$ considers number of concordant and discordant pairs (of pairs) for calculation of rank correlation. A pair of pairs $(X_i, Y_i)$, $(X_j, Y_j)$, such that $i < j$ is \textit{concordant} if the sort order matches, on $X_i, X_j$ and $Y_i, Y_j$, i.e. if $\text{sign}\big(R[X_i] - R[X_j] \big) = \text{sign}\big(R[Y_i] - R[Y_j] \big)$. Otherwise, the pair is discordant. Then $\tau$ is calculated as
\begin{equation}
    \tau = \frac{n_C - n_D}{\frac{1}{2} n (n - 1)}
\end{equation}
where $n_C$ is the number of concordant pairs, and $n_D$ is the number of discordant pairs. The value ranges from $-1$ (perfect inversion) to $+1$ (perfect agreement).

\section{Dataset Details}
\label{app:dataset}
The axes of evaluation for each dataset and number of raters per data point are detailed in Table~\ref{tab:datasets}.

\begin{table}[H]
\centering
\caption{Datasets, evaluation axes, and number of raters per datapoint.}
\label{tab:datasets}
\begin{tabular}{l p{0.55\textwidth} c}
\hline
\textbf{Dataset} & \textbf{Axes of Evaluation} & \textbf{Raters per datapoint} \\
\hline
SummEval & Coherence, Consistency, Fluency, Relevance & 8 \\
HANNA & Relevance, Coherence, Empathy, Surprise, Engagement, Complexity & 3 \\
MedVAL & Safety & 1--3 \\
MSLR & Fluency, Population, Intervention, Outcome & 1--2 \\
\hline
\end{tabular}
\label{tab:datasets}
\end{table}

\section{Rank Estimation Error Derivation}
\label{app:rank_estimation_error_connection}

The performance of metric matching relies on the human-model estimation error rank of $S^{*}$, which is, by definition the subset with rank $1$ in the inter-model estimation error. We develop the intuition more formally in this section. 

As in the main paper, consider candidate subsets $S_1,\dots, S_C \sim \mathcal{X}$. Let $\epsilon_j = \vert \widehat{\rho}_{S_j} - \rho \vert$ be the human-model estimation error for subset $S_j$, $j = 1,\dots, C$. Similarly, let $\epsilon_j^{\text{IM}} = \vert \widehat{\rho}^{\text{IM}}_{S_j} - \rho^{\text{IM}} \vert$. 
For initial understanding, consider the best-case scenario, when the rank with respect to human-model estimation error of each subset exactly matches the rank with respect to inter-model estimation error of each subset. For now, we are eliding the probabilistic component and simply assuming that this holds for all subsets $S\subseteq \mathcal{X}$ of size $b$. Formally, $R[\epsilon_j] = R[\epsilon^{\text{IM}}_j]$. This ensures that when we choose the subset $S^{*}$ with the smallest difference $\vert\widehat{\rho}_{S^{*}}^{\text{IM}} - \rho^{\text{IM}}\vert$, we choose the optimal subset w.r.t. human-model metric as well, i.e.
\begin{align*}
    S^{*} &= \arg\min_{S\subseteq X, \vert S \vert = b} \vert\widehat{\rho}_{S}^{\text{IM}} - \rho^{\text{IM}}\vert \\
    &\equiv \arg\min_{S\subseteq X, \vert S \vert = b}\vert\widehat{\rho}_{S} - \rho\vert
\end{align*}
This is an extremely strong condition and likely never satisfied in practice. Furthermore, we don't even need this to hold in order for metric matching to select the optimal subset with respect to the human-model metric. We solely require that the \textit{top} rank matches, i.e. that $R[\epsilon_j] = R[\epsilon^{\text{IM}}_j]$ for any $j = 1,\dots, C$ such that the rank equals $1$. In this case, metric matching will always choose the optimal subset (over possible candidate subsets $S_1,\dots, S_C$, not over all subsets). 

Again, in practice, this is unlikely to occur but it gives some intuition for how to understand when metric matching should work (and in particular, when metric matching should beat random selection). On a high-level, when the rank relationship between human-model estimation error and inter-model estimation error with respect to subsets of size $b$ is as close as possible to that condition, we can expect to perform well. We formalize this more below by comparison to random.

In particular, consider the \textit{expected} micro win-rate of our method. This depends both on the expectation of estimation errors of $C$ random subsets, and the estimation error of a $C+1$th subset, selected by random selection. We provide a brief derivation that shows: when the expected rank $R[\epsilon_{j^{*}}]$ of the human-model estimation error of the \textit{best} subset (rank $1$) w.r.t. the inter-model estimation is less than $\frac{C+1}{2}$, we beat random the majority of the time. Furthermore, our expected micro win-rate is linear in the expected rank $R[\epsilon_{j^{*}}]$.

\begin{lemma}
    Given the notation introduced in this section, the expected micro win rate:
    \begin{align*}
        \mathbb{E}\big[\mathbbm{1}[\epsilon_{j^{*}} < \epsilon^{\text{rand}}]\big] &= \Pr[\epsilon_{j^{*}} < \epsilon^{\text{rand}}] \\
        &= 1 - \frac{\mathbb{E}[R[\epsilon_{j^{*}}]]}{C+1} 
    \end{align*}
\end{lemma}

This admits a straightforward derivation. Given budget $b$, randomly sampled candidate subsets $S_1,\dots, S_C$ from which $S^{*} = S_{j^{*}}$ is selected by the condition that $R[\epsilon^{\text{IM}}_{j^{*}}] = 1$, consider the rank of subset $j^{*}$ w.r.t. the human-model estimation error over subsets $S_1,\dots, S_C$ \textbf{and} the subset selected by random selection $S_r$. Since all subsets are selected i.i.d. from the same distribution over $\mathcal{X}^b$, we can concretely analyze the chance that $S_r$ has a better rank than $S_{j^{*}}$ in the human-model estimation error. 
\begin{align*}
    \Pr[\epsilon_{j^{*}} < \epsilon^{\text{rand}}] &= \sum_{j=1}^C \Pr[R[\epsilon^{\text{rand}}] > j, R[\epsilon_{j^{*}}] = j] \\
    &= \sum_{j=1}^C\Pr[R[\epsilon^{\text{rand}}] > j \mid R[\epsilon_{j^{*}}] = j] \cdot \Pr[R[\epsilon_{j^{*}}] = j] \\
    &= \sum_{j=1}^C\Pr[R[\epsilon^{\text{rand}}] > j] \cdot \Pr[R[\epsilon_{j^{*}}] = j] \\
    &= \sum_{j=1}^C \Big(1 - \frac{j}{C+1}\Big) \Pr[R[\epsilon_{j^{*}}] = j] \\
    &= 1 - \frac{\mathbb{E}[R[\epsilon_{j^{*}}]]}{C+1} \\
\end{align*}
The third to fourth line in the above derivation follows from the formula for the random variable $\epsilon^{\text{rand}}$ having greater than rank $j$ out of $C+1$ variables (estimation errors $\epsilon_j$ on each of $S_1,\dots, S_C$ and $S_r$).

\begin{condition}[Expected rank condition]
\label{cond:expectedrank}
    $\mathbb{E}\big[R[\epsilon_{j^{*}}]\big] < \frac{C + 1}{2}$
\end{condition}

\begin{theorem}
    If the expected rank condition (Condition \ref{cond:expectedrank}) is satisfied, then the expected micro win-rate over random selection is greater than $0.5$, i.e. metric matching beats random the majority of the time.
\end{theorem}

\begin{figure}[H]
    \centering
    \includegraphics[width=\linewidth]{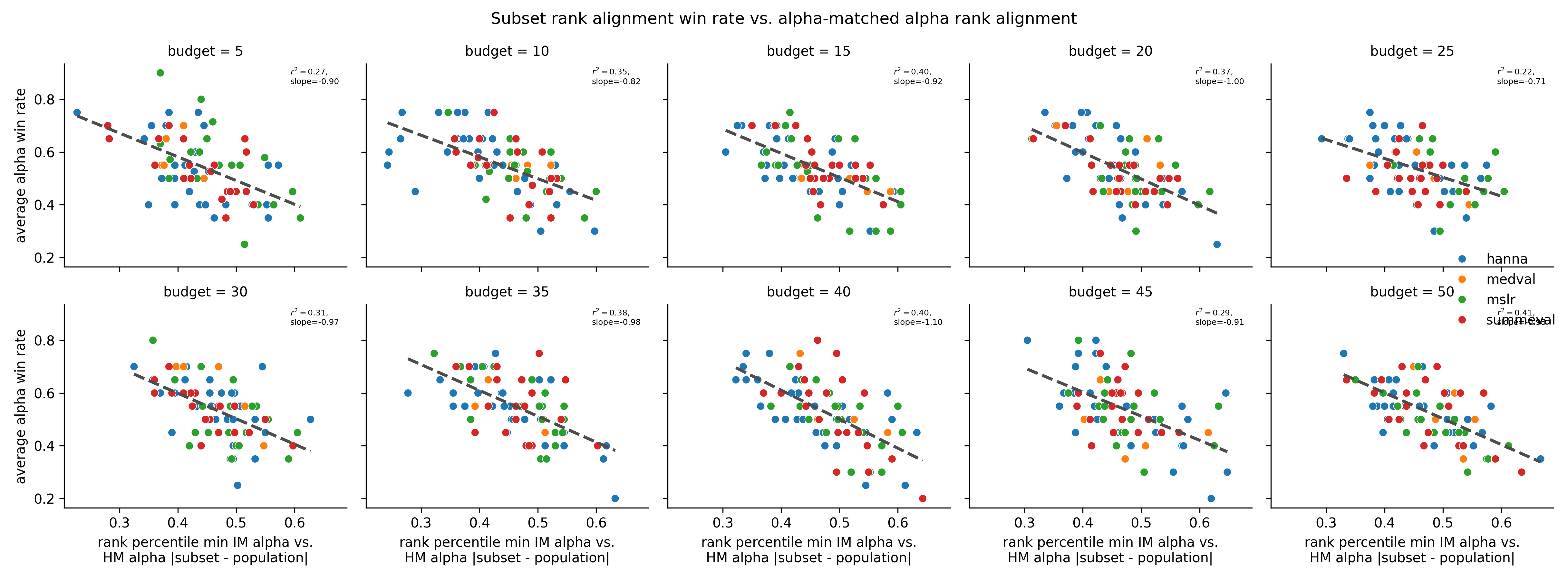}
    \caption{Empirical micro win-rate versus empirical average human-model estimation error rank percentile, $\displaystyle \frac{\sum_{t=1}^N R[\epsilon_{j^{*(t)}}]}{C+1}$, over $N = 40$ trials in metric-matched $\alpha$ estimation. As expected, the relationship is linear across budgets.}
    \label{app:fig:win_rate_rank_alpha}
\end{figure}

\begin{figure}[H]
    \centering
    \includegraphics[width=\linewidth]{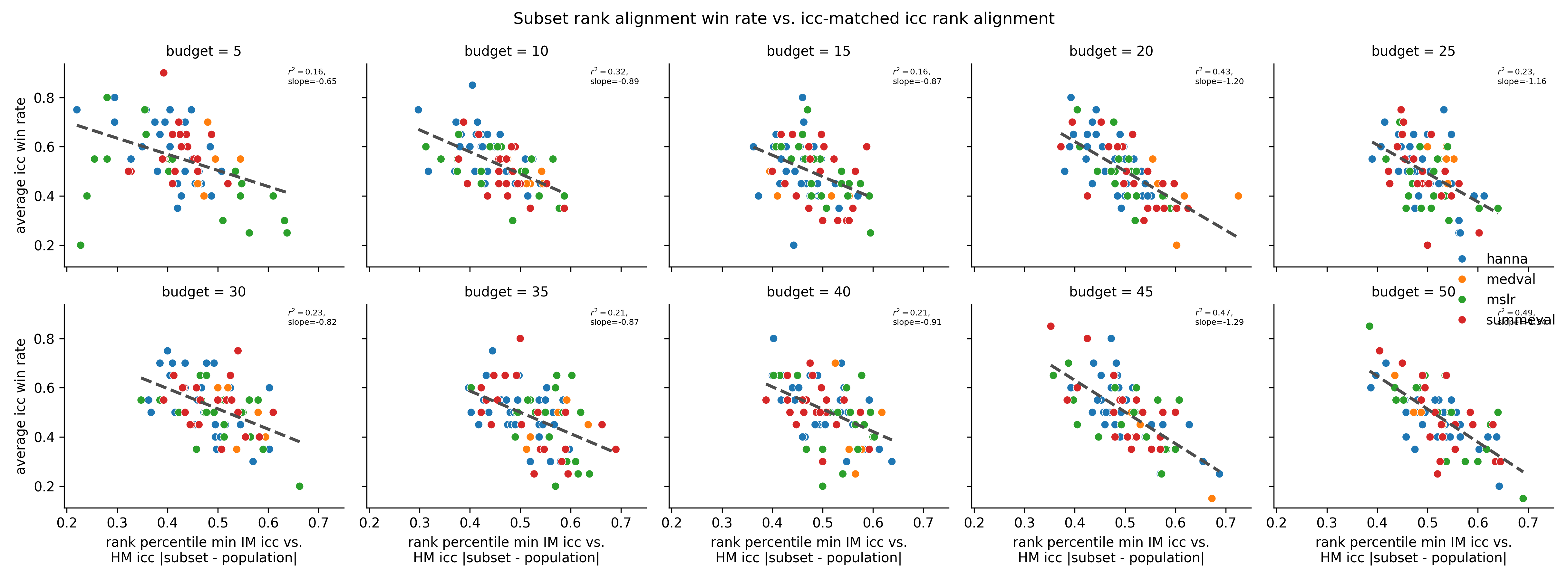}
    \caption{Observed micro win-rate versus observed average human-model estimation error rank percentile, $\displaystyle \frac{\sum_{t=1}^N R[\epsilon_{j^{*(t)}}]}{C+1}$, in metric-matched ICC estimation. As expected, the relationship is linear across budgets.}
    \label{app:fig:win_rate_rank_icc}
\end{figure}

This follows directly from the lemma. We include empirical results validating the theory in Figures~\ref{app:fig:win_rate_rank_alpha} and~\ref{app:fig:win_rate_rank_icc}.

\section{Additional Baseline Information and Comparison}
\label{app:baseline_implementation}
We present win rates for stratified sampling as well as random sampling with bias correction. For stratified sampling, the average macro win rate across metrics for estimation error is 0.794 and across threshold classification is 0.637.
\label{subsec:strat}
\begin{table}[H]
\centering
\caption{Stratified Metric Matching Performance: Estimation Error and Threshold Classification}
\begin{tabular}{c|cccc|cccc}
\toprule
& \multicolumn{4}{c|}{\textit{Estimation Error (Macro)}} & \multicolumn{4}{c}{\textit{Estimation Error (Micro)}} \\
\cmidrule(lr){2-5} \cmidrule(lr){6-9}
Budget & $\alpha$ & ICC & $\rho$ & $\tau$ & $\alpha$ & ICC & $\rho$ & $\tau$ \\
\midrule
5  & 0.827 & 0.933 & 0.947 & 0.867 & 0.563 & 0.624 & 0.597 & 0.595 \\
10 & 0.760 & 0.960 & 0.947 & 0.933 & 0.546 & 0.609 & 0.602 & 0.588 \\
15 & 0.733 & 0.907 & 0.853 & 0.880 & 0.542 & 0.595 & 0.585 & 0.569 \\
20 & 0.720 & 0.880 & 0.867 & 0.813 & 0.515 & 0.560 & 0.582 & 0.565 \\
25 & 0.693 & 0.907 & 0.813 & 0.840 & 0.514 & 0.553 & 0.582 & 0.575 \\
30 & 0.733 & 0.813 & 0.800 & 0.787 & 0.515 & 0.561 & 0.579 & 0.563 \\
35 & 0.707 & 0.787 & 0.827 & 0.787 & 0.517 & 0.543 & 0.576 & 0.544 \\
40 & 0.680 & 0.747 & 0.720 & 0.707 & 0.506 & 0.529 & 0.565 & 0.534 \\
45 & 0.640 & 0.667 & 0.747 & 0.707 & 0.528 & 0.519 & 0.565 & 0.539 \\
50 & 0.667 & 0.653 & 0.773 & 0.733 & 0.513 & 0.520 & 0.575 & 0.547 \\
\midrule
Avg & 0.716 & 0.825 & 0.829 & 0.805 & 0.526 & 0.561 & 0.581 & 0.562 \\
\bottomrule
\end{tabular}
\label{tab:variance_matching_combined}
\end{table}

\begin{table}[H]
\centering
\caption{Stratified Metric Matching Performance: Classification Win Rate (Macro and Micro)}
\begin{tabular}{c|cccc|cccc}
\toprule
& \multicolumn{4}{c|}{\textit{Classification (Macro)}} & \multicolumn{4}{c}{\textit{Classification (Micro)}} \\
\cmidrule(lr){2-5} \cmidrule(lr){6-9}
Budget & $\alpha$ & ICC & $\rho$ & $\tau$ & $\alpha$ & ICC & $\rho$ & $\tau$ \\
\midrule
5  & 0.688 & 0.585 & 0.784 & 0.824 & 0.575 & 0.554 & 0.629 & 0.644 \\
10 & 0.655 & 0.611 & 0.719 & 0.717 & 0.545 & 0.575 & 0.611 & 0.628 \\
15 & 0.665 & 0.623 & 0.701 & 0.676 & 0.557 & 0.585 & 0.597 & 0.591 \\
20 & 0.571 & 0.625 & 0.703 & 0.619 & 0.534 & 0.564 & 0.603 & 0.554 \\
25 & 0.676 & 0.611 & 0.685 & 0.738 & 0.562 & 0.571 & 0.598 & 0.595 \\
30 & 0.598 & 0.620 & 0.713 & 0.697 & 0.547 & 0.573 & 0.613 & 0.575 \\
35 & 0.623 & 0.625 & 0.679 & 0.603 & 0.530 & 0.571 & 0.595 & 0.554 \\
40 & 0.527 & 0.655 & 0.672 & 0.423 & 0.493 & 0.574 & 0.573 & 0.469 \\
45 & 0.532 & 0.630 & 0.604 & 0.653 & 0.510 & 0.574 & 0.556 & 0.591 \\
50 & 0.484 & 0.616 & 0.577 & 0.479 & 0.483 & 0.564 & 0.559 & 0.486 \\
\midrule
Avg & 0.602 & 0.620 & 0.684 & 0.643 & 0.534 & 0.570 & 0.593 & 0.569 \\
\bottomrule
\end{tabular}
\label{tab:variance_matching_combined}
\end{table}

For random selection with bias correction, the average macro win rate across metrics for estimation error is 0.863 and across threshold classification is 0.392. 

\begin{table}[H]
\centering
\caption{Random with Bias Correction Metric Matching Performance: Estimation Error (Macro and Micro)}
\begin{tabular}{c|cccc|cccc}
\toprule
& \multicolumn{4}{c|}{\textit{Estimation Error (Macro)}} & \multicolumn{4}{c}{\textit{Estimation Error (Micro)}} \\
\cmidrule(lr){2-5} \cmidrule(lr){6-9}
Budget & $\alpha$ & ICC & $\rho$ & $\tau$ & $\alpha$ & ICC & $\rho$ & $\tau$ \\
\midrule
5  & 0.880 & 0.960 & 0.853 & 0.867 & 0.540 & 0.595 & 0.547 & 0.562 \\
10 & 0.787 & 1.000 & 0.947 & 0.987 & 0.537 & 0.569 & 0.572 & 0.591 \\
15 & 0.773 & 0.973 & 0.947 & 0.933 & 0.540 & 0.579 & 0.572 & 0.575 \\
20 & 0.707 & 0.907 & 0.973 & 0.947 & 0.532 & 0.554 & 0.583 & 0.588 \\
25 & 0.720 & 0.880 & 0.933 & 0.960 & 0.545 & 0.558 & 0.590 & 0.589 \\
30 & 0.733 & 0.840 & 0.920 & 0.867 & 0.532 & 0.546 & 0.598 & 0.569 \\
35 & 0.733 & 0.813 & 0.920 & 0.827 & 0.543 & 0.536 & 0.613 & 0.581 \\
40 & 0.747 & 0.827 & 0.947 & 0.867 & 0.548 & 0.539 & 0.602 & 0.559 \\
45 & 0.760 & 0.800 & 0.947 & 0.813 & 0.563 & 0.551 & 0.622 & 0.569 \\
50 & 0.733 & 0.733 & 0.947 & 0.827 & 0.569 & 0.540 & 0.631 & 0.579 \\
\midrule
Avg & 0.757 & 0.873 & 0.933 & 0.889 & 0.545 & 0.557 & 0.593 & 0.576 \\
\bottomrule
\end{tabular}
\label{tab:variance_matching_combined}
\end{table}

\begin{table}[H]
\centering
\caption{Random with Bias Correction Metric Matching Performance: Thresholding (Macro and Micro)}
\begin{tabular}{c|cccc|cccc}
\toprule
& \multicolumn{4}{c|}{\textit{Thresholding (Macro)}} & \multicolumn{4}{c}{\textit{Thresholding (Micro)}} \\
\cmidrule(lr){2-5} \cmidrule(lr){6-9}
Budget & $\alpha$ & ICC & $\rho$ & $\tau$ & $\alpha$ & ICC & $\rho$ & $\tau$ \\
\midrule
5  & 0.398 & 0.537 & 0.363 & 0.431 & 0.443 & 0.531 & 0.429 & 0.463 \\
10 & 0.289 & 0.484 & 0.387 & 0.611 & 0.372 & 0.526 & 0.443 & 0.542 \\
15 & 0.270 & 0.505 & 0.309 & 0.548 & 0.361 & 0.533 & 0.385 & 0.511 \\
20 & 0.246 & 0.506 & 0.285 & 0.455 & 0.310 & 0.522 & 0.374 & 0.482 \\
25 & 0.204 & 0.496 & 0.306 & 0.614 & 0.318 & 0.533 & 0.343 & 0.533 \\
30 & 0.165 & 0.505 & 0.336 & 0.521 & 0.275 & 0.527 & 0.382 & 0.522 \\
35 & 0.165 & 0.518 & 0.290 & 0.569 & 0.275 & 0.534 & 0.354 & 0.551 \\
40 & 0.141 & 0.520 & 0.342 & 0.419 & 0.235 & 0.538 & 0.369 & 0.517 \\
45 & 0.158 & 0.539 & 0.271 & 0.504 & 0.213 & 0.557 & 0.327 & 0.565 \\
50 & 0.167 & 0.507 & 0.259 & 0.540 & 0.227 & 0.542 & 0.367 & 0.526 \\
\midrule
Avg & 0.220 & 0.512 & 0.315 & 0.521 & 0.303 & 0.534 & 0.377 & 0.521 \\
\bottomrule
\end{tabular}
\label{tab:variance_matching_combined}
\end{table}

\section{Ablation Studies}
\subsection{Average Aggregation}
\label{app:average_then_pairwise}

\begin{figure}[H]
    \centering
    \begin{subfigure}{0.45\textwidth}
        \centering
        \includegraphics[width=\linewidth]{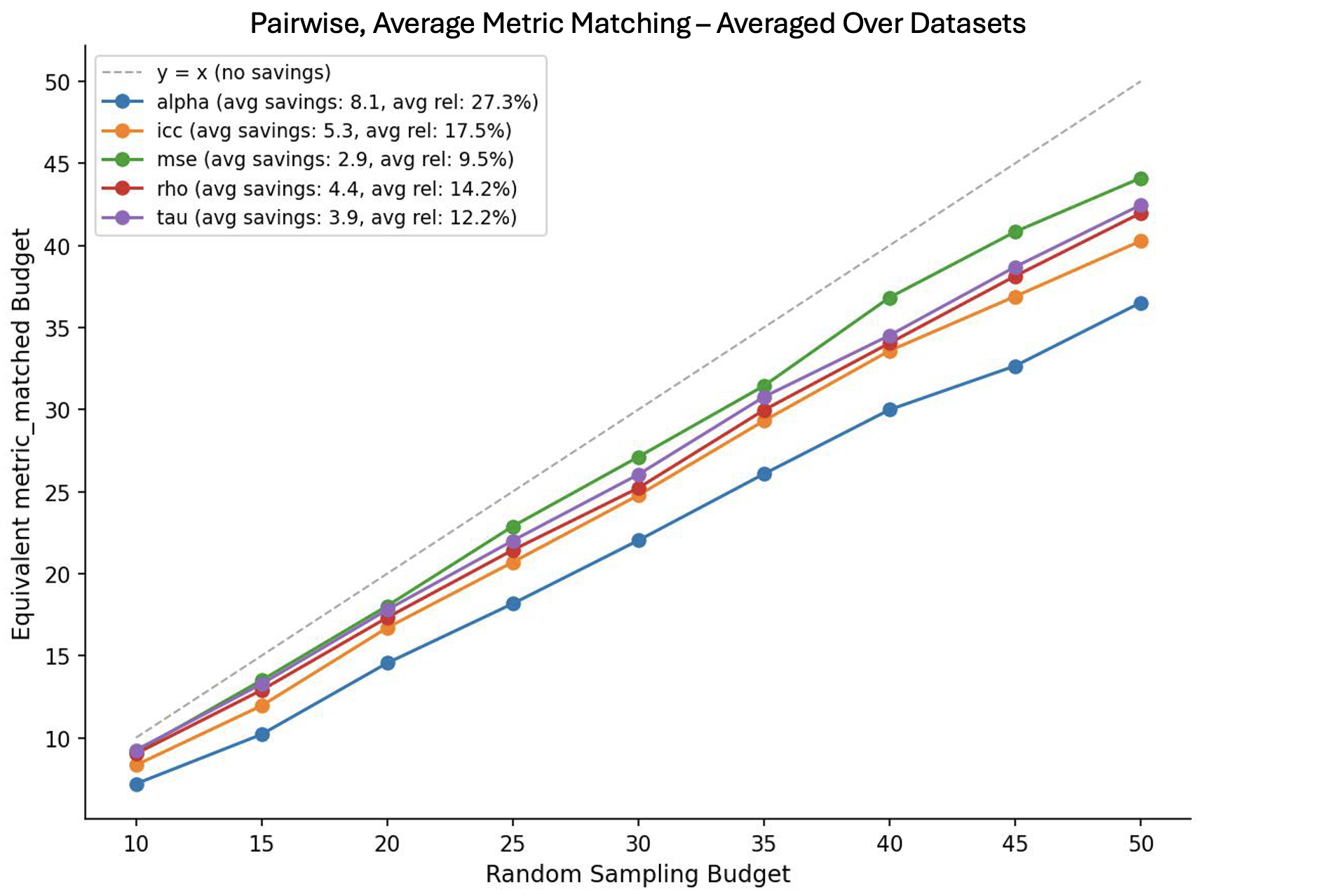}
        \caption{}
    \end{subfigure}
    \hfill
    \begin{subfigure}{0.45\textwidth}
        \centering
        \includegraphics[width=\linewidth]{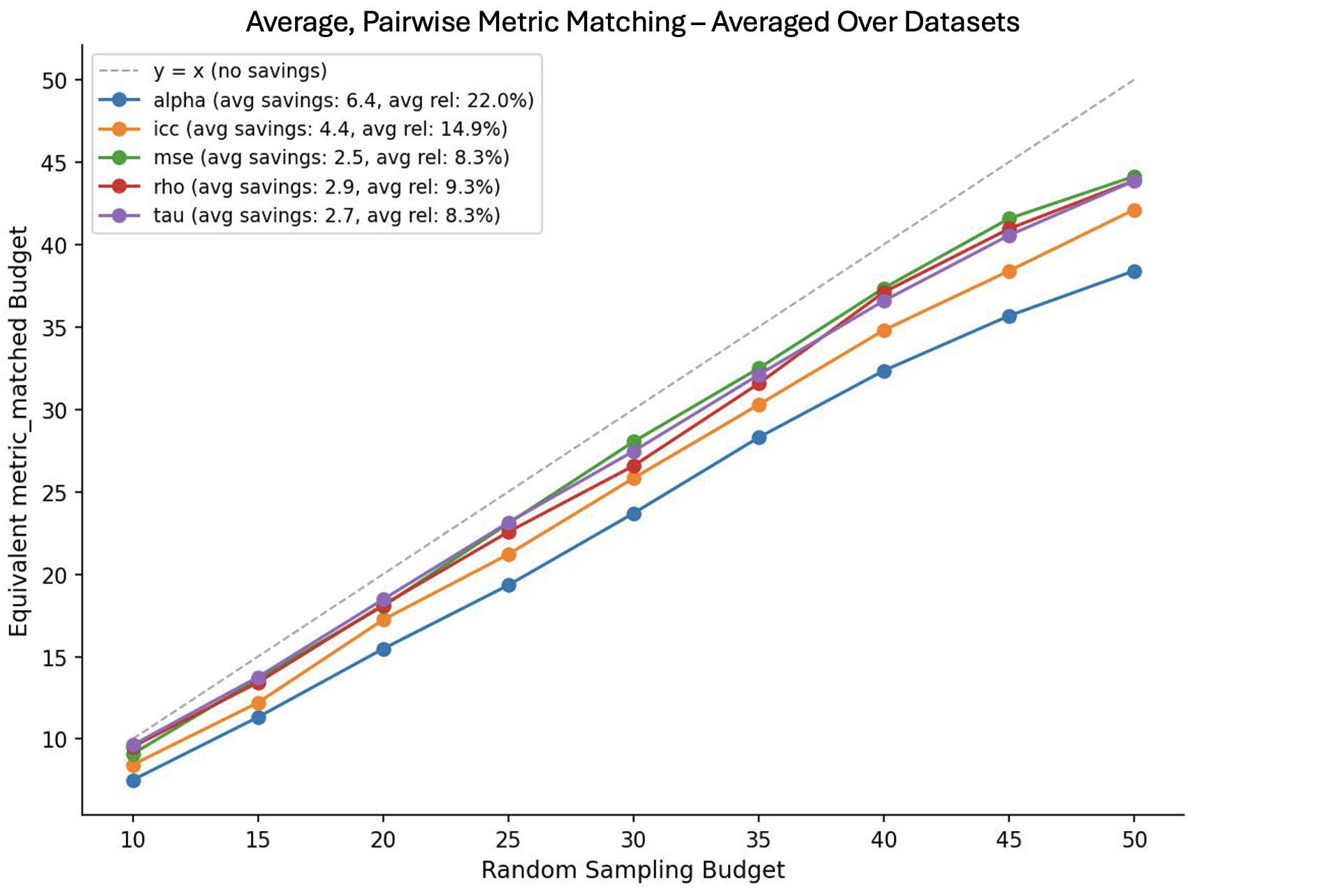}
        \caption{}
    \end{subfigure}

    \caption{Aggregation approaches impact effectiveness of \textbf{Metric Match}, with the optimal approach being to calculate the reliability of our target model with each ensemble member individually and average these results (a) as opposed to averaging the ensemble labels and calculating the reliability coefficient with a single "averaged" ensemble point (b).}
    \label{fig:avg_pair}
\end{figure}

We show that the way we aggregate model information impacts the performance of our algorithm. Given a model suite $\mathcal{M} = \{M_1, M_2, \ldots, M_n\}$ and a target model $T \in \mathcal{M}$, we present an alternative aggregation approach in Figure~\ref{fig:avg_pair}b. We calculate the target metric by
$\hat{\rho} = f\left(T, \frac{1}{|\mathcal{M}|}\sum_{M_i \in \mathcal{M}} M_i\right)$. This essentially constructs an average score function from the scorers in $\mathcal{M}$, $M^{\text{avg}}: \mathcal{X} \rightarrow\mathcal{Y}$ s.t. $M^{\text{avg}}(x) = \frac{1}{|\mathcal{M}|}\sum_{M'\in\mathcal{M}} M'(x)$. We posit that this fails to capture all necessary signal from individual judges, thus causing the alternative aggregation method that we employ in the main body to perform better.

Estimation error win rates as shown in Table~\ref{tab:metric_matching_combined} are higher for this aggregation method, indicating that although the gains are less significant, they are equally if not more reliable.


\begin{table}[h]
\centering
\caption{Average-Pairwise Metric Matching Performance: Estimation Error and Threshold Classification Macro Win Rate}
\begin{tabular}{c|cccc|cccc}
\hline
 & \multicolumn{4}{c|}{\textit{Estimation Error (Win Rate)}} & \multicolumn{4}{c}{\textit{Threshold Classification (Win Rate)}} \\
Budget & $\alpha$ & ICC & $\tau$ & $\rho$ & $\alpha$ & ICC & $\tau$ & $\rho$ \\
\hline
5  & 0.867 & 0.760 & 0.587 & 0.507 & 0.631 & 0.625 & 0.797 & 0.671 \\
10 & 0.920 & 0.840 & 0.760 & 0.667 & 0.623 & 0.667 & 0.739 & 0.611 \\
15 & 0.893 & 0.787 & 0.773 & 0.733 & 0.750 & 0.653 & 0.661 & 0.791 \\
20 & 0.867 & 0.733 & 0.693 & 0.680 & 0.644 & 0.612 & 0.571 & 0.714 \\
25 & 0.853 & 0.733 & 0.680 & 0.707 & 0.667 & 0.585 & 0.630 & 0.774 \\
30 & 0.773 & 0.720 & 0.733 & 0.773 & 0.795 & 0.516 & 0.730 & 0.655 \\
35 & 0.840 & 0.680 & 0.627 & 0.733 & 0.675 & 0.477 & 0.649 & 0.623 \\
40 & 0.760 & 0.720 & 0.640 & 0.587 & 0.694 & 0.550 & 0.630 & 0.700 \\
45 & 0.853 & 0.747 & 0.640 & 0.653 & 0.611 & 0.567 & 0.600 & 0.660 \\
50 & 0.880 & 0.800 & 0.720 & 0.720 & 0.581 & 0.613 & 0.556 & 0.651 \\
\hline
Avg & 0.851 & 0.752 & 0.685 & 0.676 & 0.667 & 0.587 & 0.656 & 0.685 \\
\bottomrule
\end{tabular}
\label{tab:metric_matching_combined}
\end{table}

\subsection{Number of Candidate Subsets $C$}
\label{app:subsec:n_candidates}
\begin{figure}[H]
    \centering
    \begin{subfigure}{\textwidth}
        \subcaption{HANNA}
        \includegraphics[width=\textwidth]{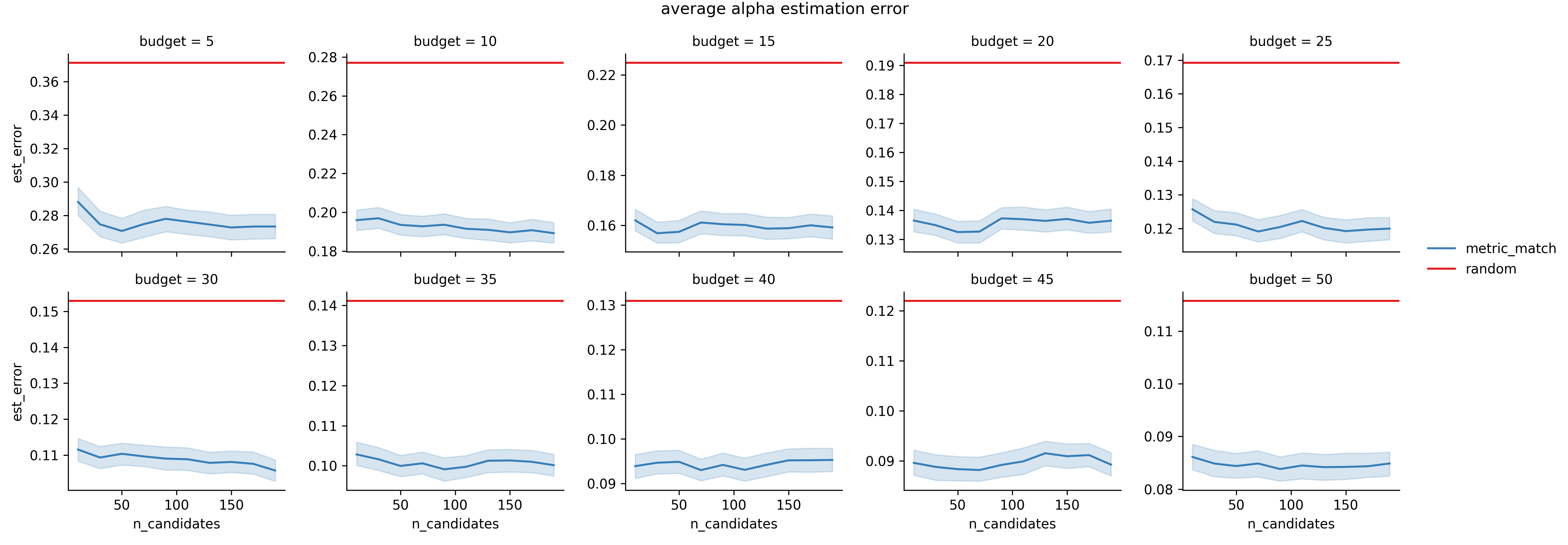}
    \end{subfigure}
    \begin{subfigure}{\textwidth}
        \subcaption{MedVAL}
        \includegraphics[width=\textwidth]{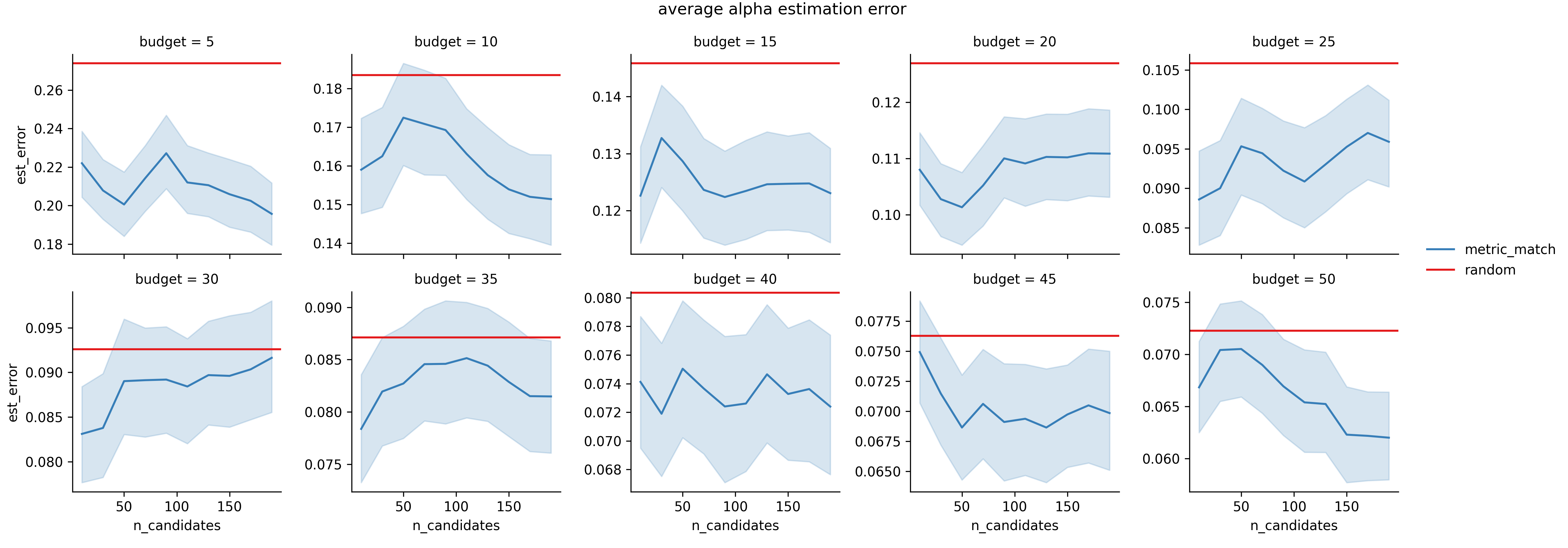}
    \end{subfigure}
    \begin{subfigure}{\textwidth}
        \subcaption{MSLR}
        \includegraphics[width=\textwidth]{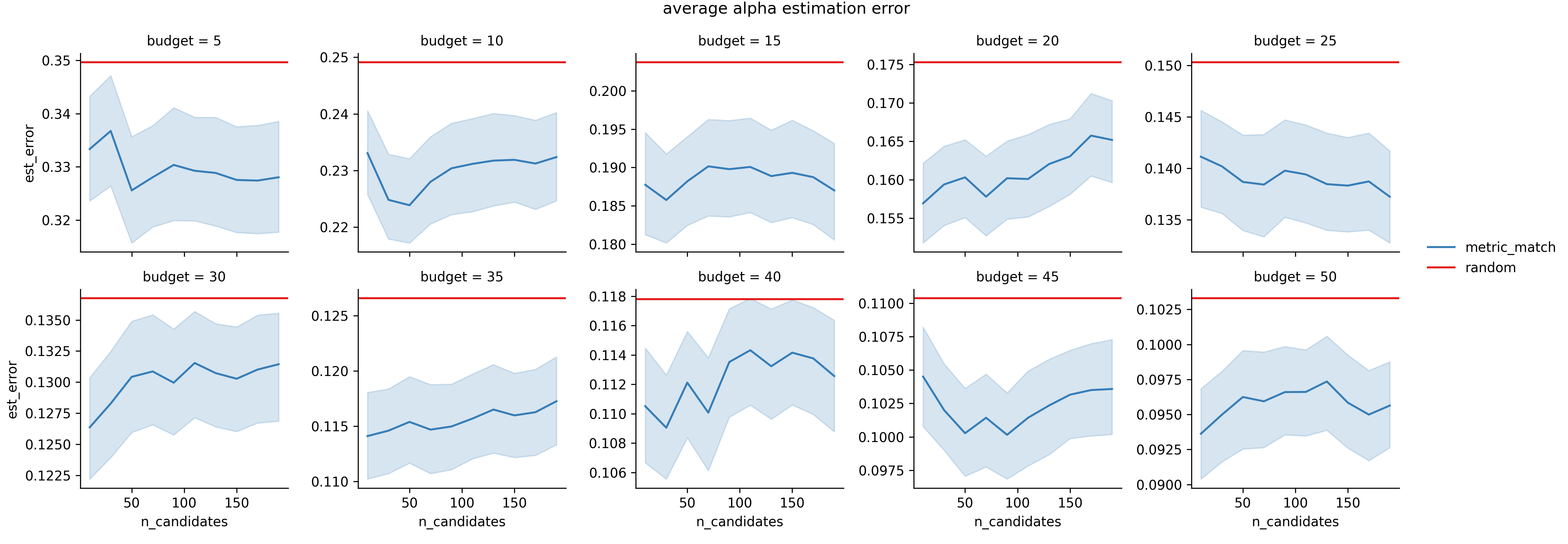}
    \end{subfigure}
    \begin{subfigure}{\textwidth}
        \subcaption{SummEval}
        \includegraphics[width=\textwidth]{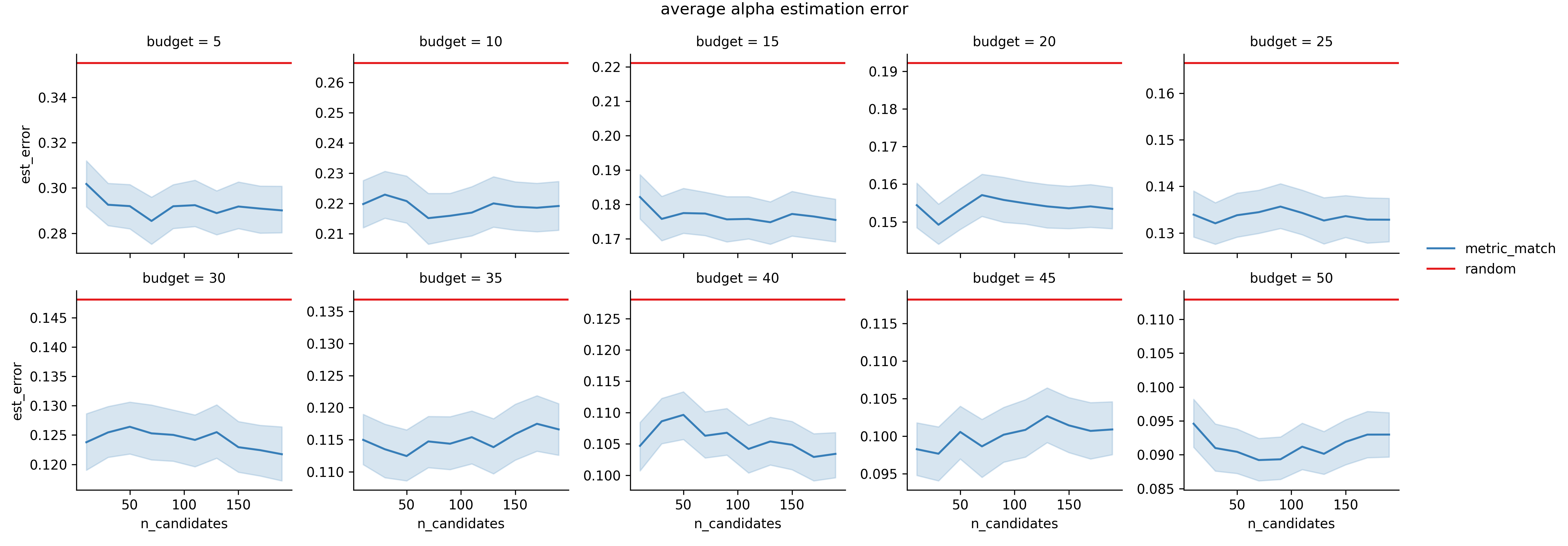}
    \end{subfigure}
    \caption{Ablation study over number of candidate subsets considered: average estimation error in $\alpha$ metric matching at each budget for each dataset for runs of metric matching with $C = 10$ up to $C=200$ candidate subsets with $N = 40$ trials each.}
    \label{app:fig:n_candidates_alpha}
\end{figure}

\begin{figure}[H]
    \centering
    \begin{subfigure}{\textwidth}
        \subcaption{HANNA}
        \includegraphics[width=\textwidth]{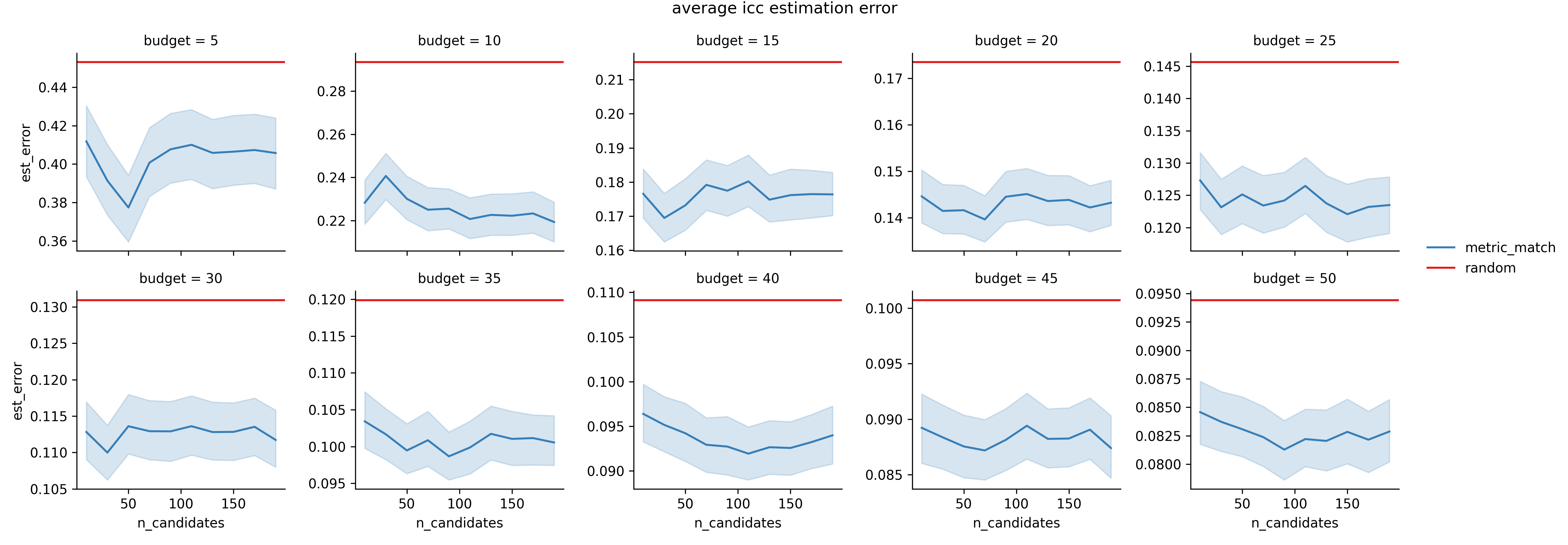}
    \end{subfigure}
    \begin{subfigure}{\textwidth}
        \subcaption{MedVAL}
        \includegraphics[width=\textwidth]{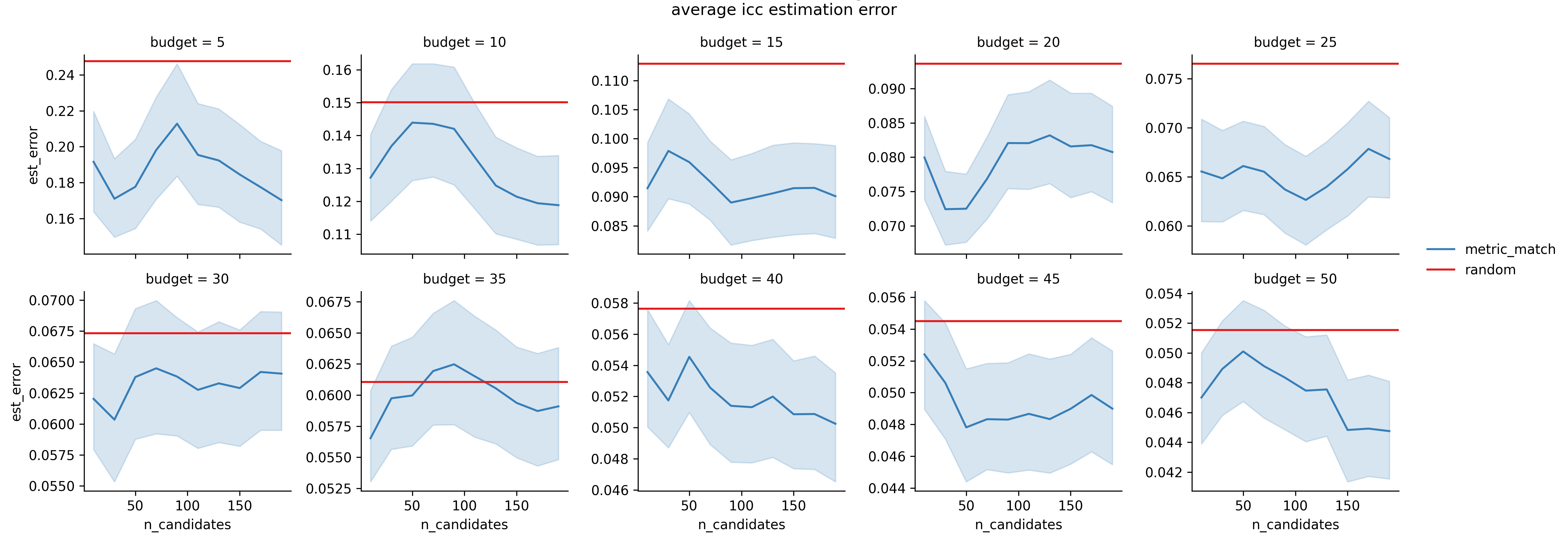}
    \end{subfigure}
    \begin{subfigure}{\textwidth}
        \subcaption{MSLR}
        \includegraphics[width=\textwidth]{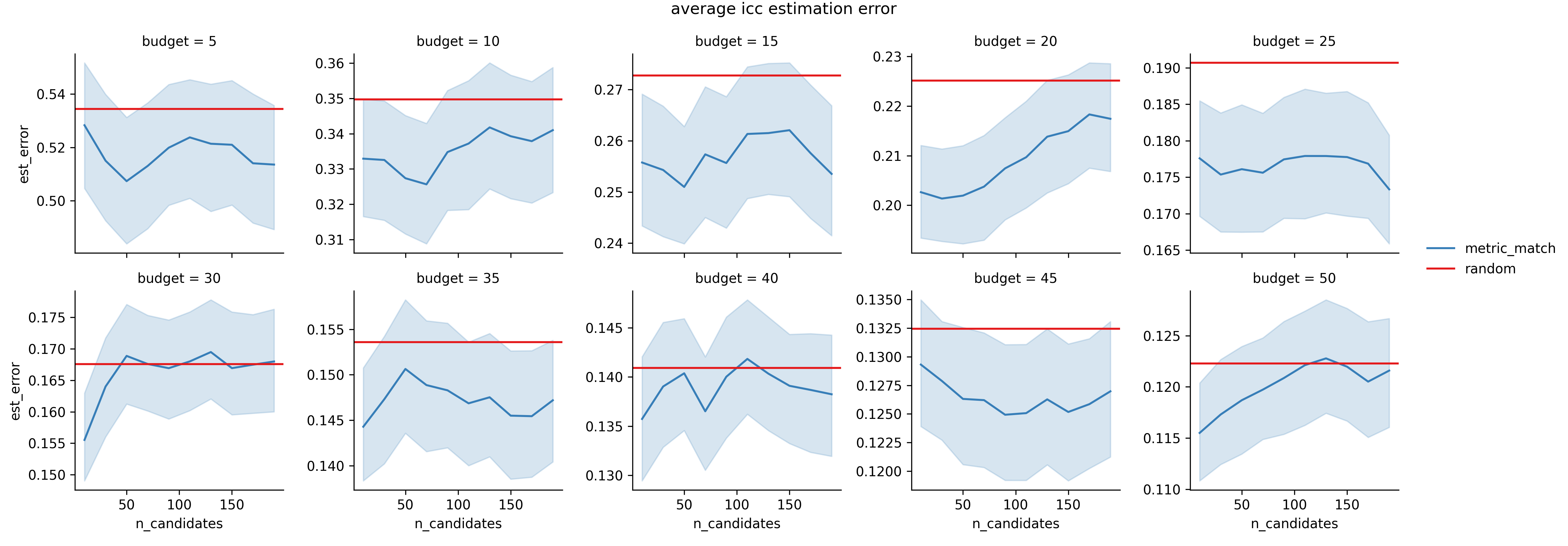}
    \end{subfigure}
    \begin{subfigure}{\textwidth}
        \subcaption{SummEval}
        \includegraphics[width=\textwidth]{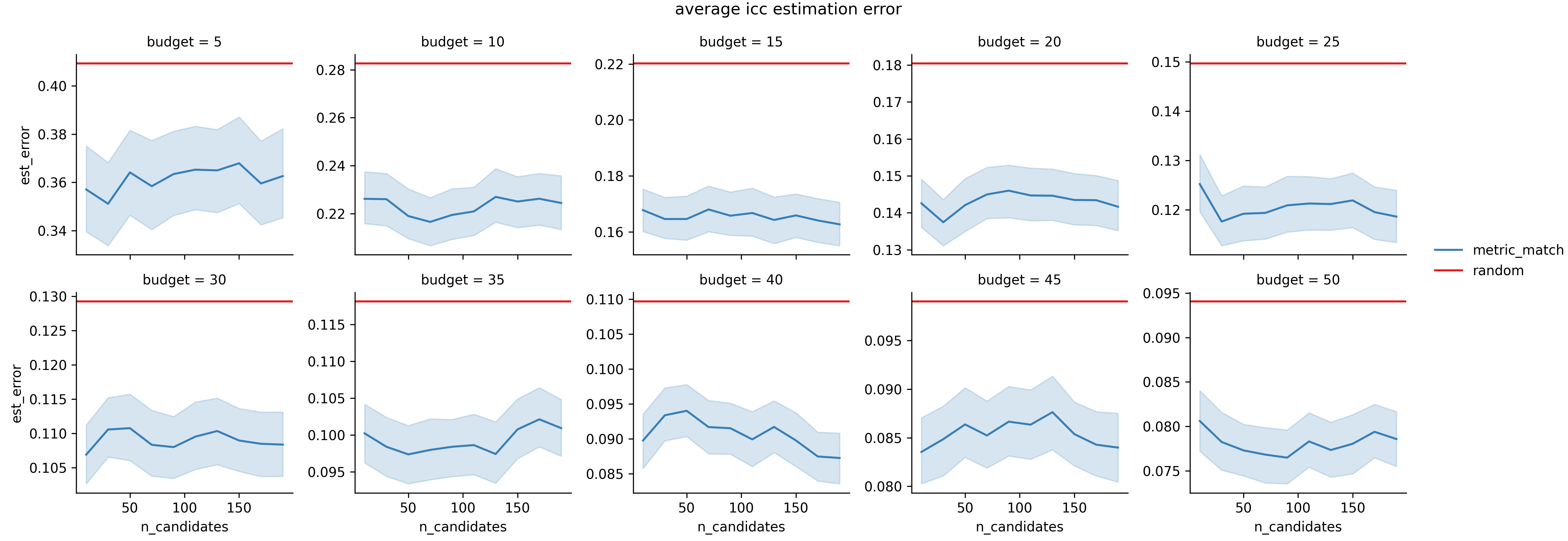}
    \end{subfigure}
    \caption{Ablation study over number of candidate subsets considered: average estimation error in ICC metric matching at each budget for each dataset for runs of metric matching with $C = 10$ up to $C=200$ candidate subsets with $N = 40$ trials each.}
    \label{app:fig:n_candidates_icc}
\end{figure}

Here, we vary the number of candidate subsets considered for variance matching between $C = 10$ to $C = 200$, and report the estimation errors with comparison to random baseline in the red horizontal line in Figure~\ref{app:fig:n_candidates_icc} for ICC and Figure~\ref{app:fig:n_candidates_alpha} for alpha estimation. We note that these values would correspond to macro-average win-rate and thus do not represent the results we would expect to see with micro-average win-rate. We do not observe an interpretable trend across datasets or budgets between macro-average and number of candidate subsets. Additionally, due to computational constraints, we only study up to $200$ candidate subsets, which may be well below the number required to observe any relationship (recall total possible subsets is $\displaystyle {|\mathcal{X}| \choose b}$).

\section{Variance Matching}
\label{subsec:var_matching}
We present an alternative algorithm as an extension to metric matching called \textit{Variance Matching}. We find that Variance Matching performs worse than metric matching on average, but in cases where metric matching performs very poorly, variance matching is more robust to these outlying cases.

\begin{algorithm}[H]
\caption{Weighted Variance Matching}
\label{alg:weighted_vm}
\DontPrintSemicolon
\KwData{Metric function $T$; MSB function $\texttt{MSB}$, MSE function $\texttt{MSE}$; weights $\alpha, \beta$; budget $b$; generated text data $X = \{x_i\}_{i=1}^n \in \mathcal{X}^n$; scores $\{y_i^{(M)}\}_{i=1}^n$, \{$y_i^{(M')}\}_{i=1}^n$ from LLM judges $M$ and $M'$, respectively; $C$, number of candidate subsets over which to search}
\KwResult{Subset $S^{*} \subseteq X$, $\lvert S^{*} \rvert = b$}
Initialize $\delta_{\min} \gets \infty$\;
Initialize $S^{*} \gets \emptyset$\;
$\rho^{\text{IM}} \gets \alpha \cdot\texttt{MSB}\big(\{y^{(M)}_{i}\}_{i=1}^n, \{y^{(M')}_{i}\}_{i=1}^n\big)+\beta\cdot\texttt{MSE}\big(\{y^{(M)}_{i}\}_{i=1}^n, \{y^{(M')}_{i}\}_{i=1}^n\big)$\;
\For{$j = 1, \dots, C$}{
    Sample $S_j = \{x_{i_1}, \dots, x_{i_b}\}$ uniformly from $X$ without replacement\;
    $\widehat{\rho}_{S_j}^{\text{IM}} \gets \alpha\cdot \texttt{MSB}\big(\{y^{(M)}_{i_k}\}_{k=1}^b, \{y^{(M')}_{i_k}\}_{k=1}^b\big)+\beta\cdot \texttt{MSE}\big(\{y^{(M)}_{i_k}\}_{k=1}^b, \{y^{(M')}_{i_k}\}_{k=1}^b\big)$\;
    $\delta_j = \lvert \widehat{\rho}_{S_j}^{\text{IM}} - \rho^{\text{IM}}\rvert$\;
    \If{$\delta_j < \delta_{\min}$}{
        $S^{*} \gets S_j$\;
        $\delta_{\min} \gets \delta_j$
    }
}
\Return{$S^{*}$}
\end{algorithm}

We find that the optimal hyperparameters for this mixture are $\alpha=0.9$ and $\beta=0.1$. We highlight results of this algorithm in Table~\ref{tab:variance_matching_combined}.

\begin{table}[H]
\centering
\caption{Variance Matching Performance: Estimation Error and Threshold Classification}
\begin{tabular}{c|cccc|cccc}
\toprule
& \multicolumn{4}{c|}{\textit{Estimation Error (Win Rate)}} & \multicolumn{4}{c}{\textit{Threshold Classification (Win Rate)}} \\
\cmidrule(lr){2-5} \cmidrule(lr){6-9}
Budget & $\alpha$ & ICC & $\tau$ & $\rho$ & $\alpha$ & ICC & $\tau$ & $\rho$ \\
\midrule
5  & 0.867 & 0.840 & 0.760 & 0.800 & 0.655 & 0.505 & 0.585 & 0.572 \\
10 & 0.920 & 0.947 & 0.880 & 0.880 & 0.728 & 0.539 & 0.710 & 0.653 \\
15 & 0.973 & 0.947 & 0.773 & 0.827 & 0.667 & 0.548 & 0.587 & 0.627 \\
20 & 0.933 & 0.907 & 0.827 & 0.840 & 0.506 & 0.496 & 0.333 & 0.511 \\
25 & 0.867 & 0.867 & 0.773 & 0.787 & 0.574 & 0.609 & 0.500 & 0.640 \\
30 & 0.880 & 0.947 & 0.800 & 0.840 & 0.620 & 0.583 & 0.720 & 0.640 \\
35 & 0.933 & 0.907 & 0.787 & 0.800 & 0.563 & 0.642 & 0.464 & 0.488 \\
40 & 0.760 & 0.827 & 0.707 & 0.733 & 0.466 & 0.539 & 0.452 & 0.449 \\
45 & 0.907 & 0.893 & 0.773 & 0.800 & 0.433 & 0.586 & 0.461 & 0.504 \\
50 & 0.853 & 0.867 & 0.733 & 0.760 & 0.679 & 0.522 & 0.558 & 0.442 \\
\midrule
Avg & 0.889 & 0.895 & 0.781 & 0.807 & 0.589 & 0.557 & 0.537 & 0.552 \\
\bottomrule
\end{tabular}
\label{tab:variance_matching_combined}
\end{table}

\section{Mean Squared Error}
\label{subsec:mse}
We introduce results with non-correlation-based metrics, such as Mean Squared Error (MSE). We show that variance matching as opposed to \textbf{Metric Match} on the MSE metric provides larger gains, although \textbf{Metric Match} still outperforms random for estimation error and threshold classification at the macro level for small budgets.


\begin{table}[H]
\centering
\caption{MSE Win Rates by Method: Macro and Micro Win Rates}
\begin{tabular}{c|cc|cc|cc|cc}
\toprule
& \multicolumn{4}{c|}{\textit{Estimation Error}} & \multicolumn{4}{c}{\textit{Threshold Classification}} \\
\cmidrule(lr){2-5} \cmidrule(lr){6-9}
& \multicolumn{2}{c|}{Metric Matching} & \multicolumn{2}{c|}{Variance Matching} & \multicolumn{2}{c|}{Metric Matching} & \multicolumn{2}{c}{Variance Matching} \\
\cmidrule(lr){2-3} \cmidrule(lr){4-5} \cmidrule(lr){6-7} \cmidrule(lr){8-9}
Budget & Macro & Micro & Macro & Micro & Macro & Micro & Macro & Micro \\
\midrule
5  & 0.827 & 0.475 & 0.733 & 0.481 & 0.568 & 0.481 & 0.512 & 0.511 \\
10 & 0.720 & 0.492 & 0.813 & 0.517 & 0.524 & 0.469 & 0.550 & 0.531 \\
15 & 0.693 & 0.486 & 0.800 & 0.517 & 0.565 & 0.448 & 0.676 & 0.520 \\
20 & 0.587 & 0.469 & 0.733 & 0.503 & 0.462 & 0.445 & 0.567 & 0.508 \\
25 & 0.613 & 0.459 & 0.800 & 0.512 & 0.415 & 0.439 & 0.550 & 0.505 \\
30 & 0.587 & 0.450 & 0.760 & 0.500 & 0.398 & 0.434 & 0.547 & 0.495 \\
35 & 0.613 & 0.447 & 0.827 & 0.522 & 0.387 & 0.442 & 0.498 & 0.508 \\
40 & 0.573 & 0.449 & 0.640 & 0.502 & 0.392 & 0.435 & 0.512 & 0.481 \\
45 & 0.520 & 0.442 & 0.680 & 0.499 & 0.378 & 0.504 & 0.454 & 0.497 \\
50 & 0.587 & 0.446 & 0.627 & 0.492 & 0.419 & 0.450 & 0.467 & 0.485 \\
\midrule
Avg & 0.632 & 0.462 & 0.741 & 0.504 & 0.451 & 0.455 & 0.533 & 0.505 \\
\bottomrule
\end{tabular}
\label{tab:mse_win_rate}
\end{table}

\section{Results by Dataset Axis}
Here we present disagreggated results at the dataset level in order to highlight the robustness of this method across datasets, metrics, and budgets. We highlight that poor performance for certain (metric, dataset) pairs are empirically driven by low true reliability parameters, indicating that \textbf{Metric Match} may be more susceptible than variance matching when the judge is misaligned from human constructs.

\label{subsec:not_agg}
\begin{figure}[H]
    \centering
    \begin{subfigure}{0.45\textwidth}
        \centering
        \includegraphics[width=\linewidth]{figures/annotations/budget_equiv_summary-5.jpg}

    \end{subfigure}
    \hfill
    \begin{subfigure}{0.45\textwidth}
        \centering
        \includegraphics[width=\linewidth]{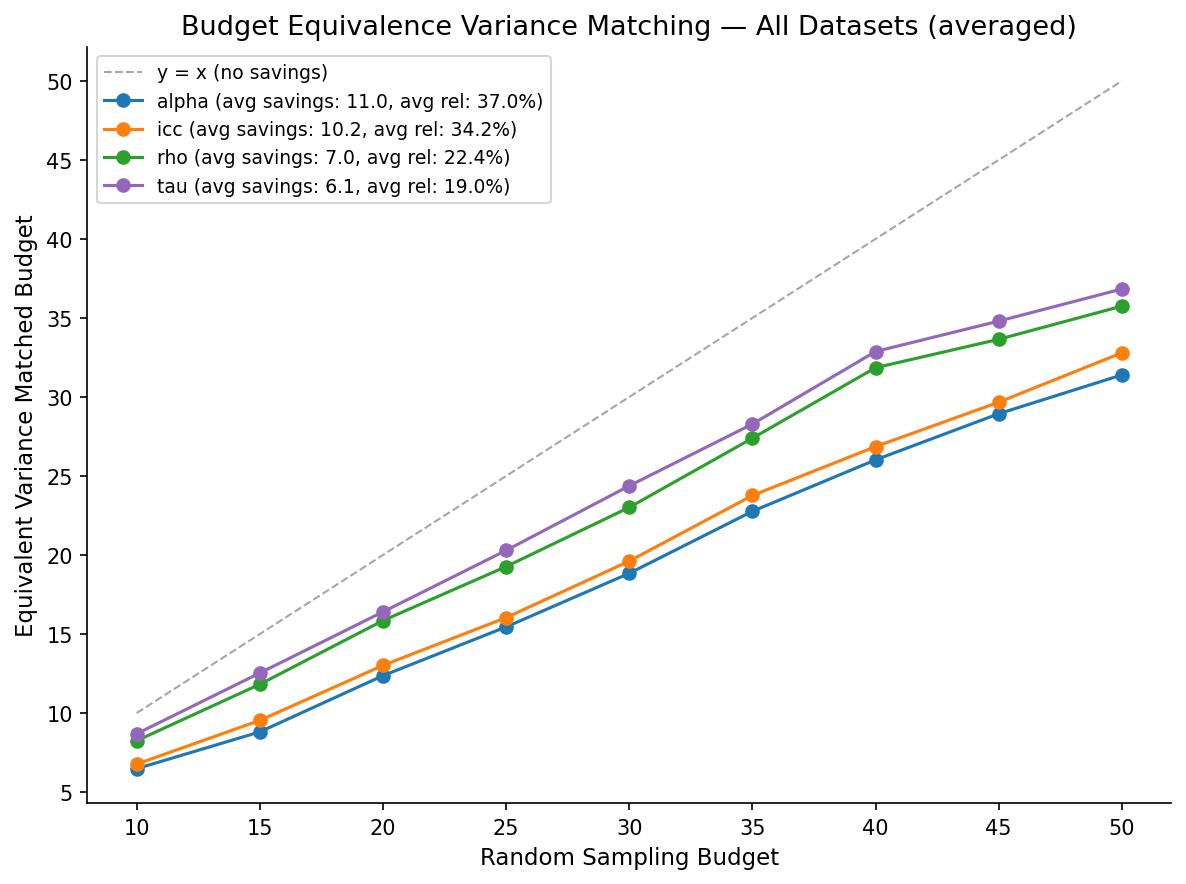}
    \end{subfigure}
        \centering
    \begin{subfigure}{0.45\textwidth}
        \centering
        \includegraphics[width=\linewidth]{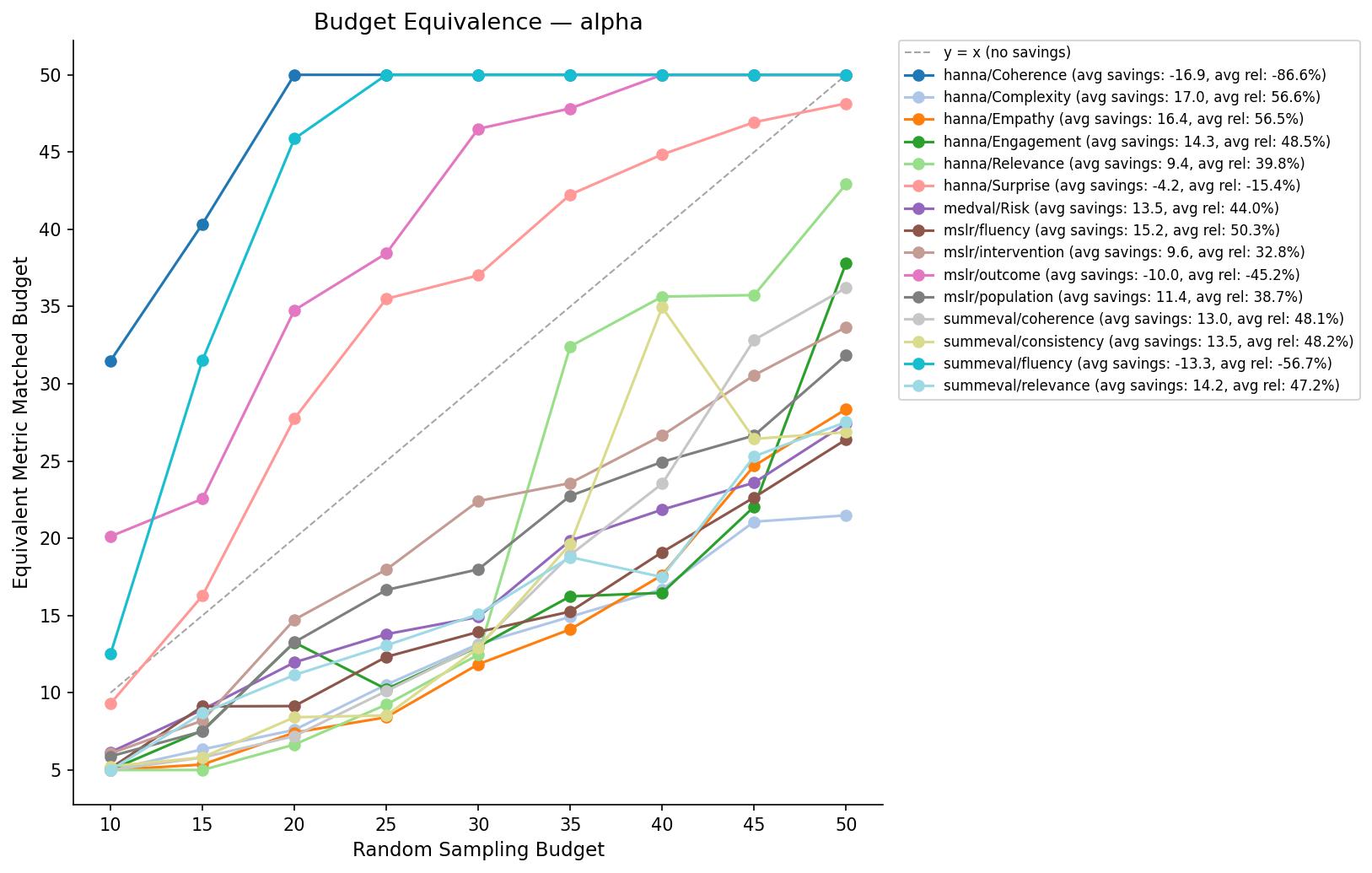}
    \end{subfigure}
    \hfill
    \begin{subfigure}{0.45\textwidth}
        \centering
        \includegraphics[width=\linewidth]{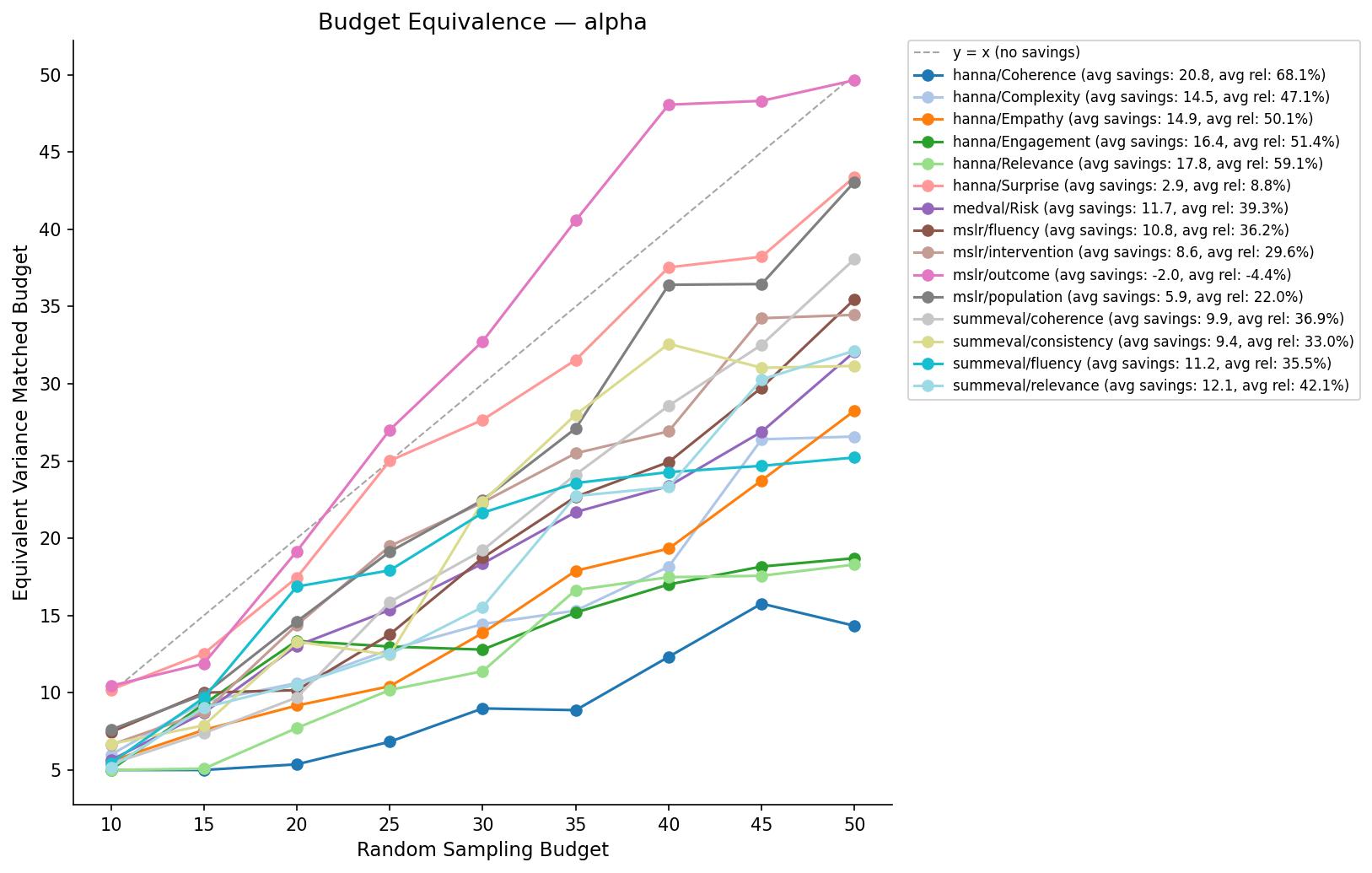}
    \end{subfigure}
        \centering
    \begin{subfigure}{0.45\textwidth}
        \centering
        \includegraphics[width=\linewidth]{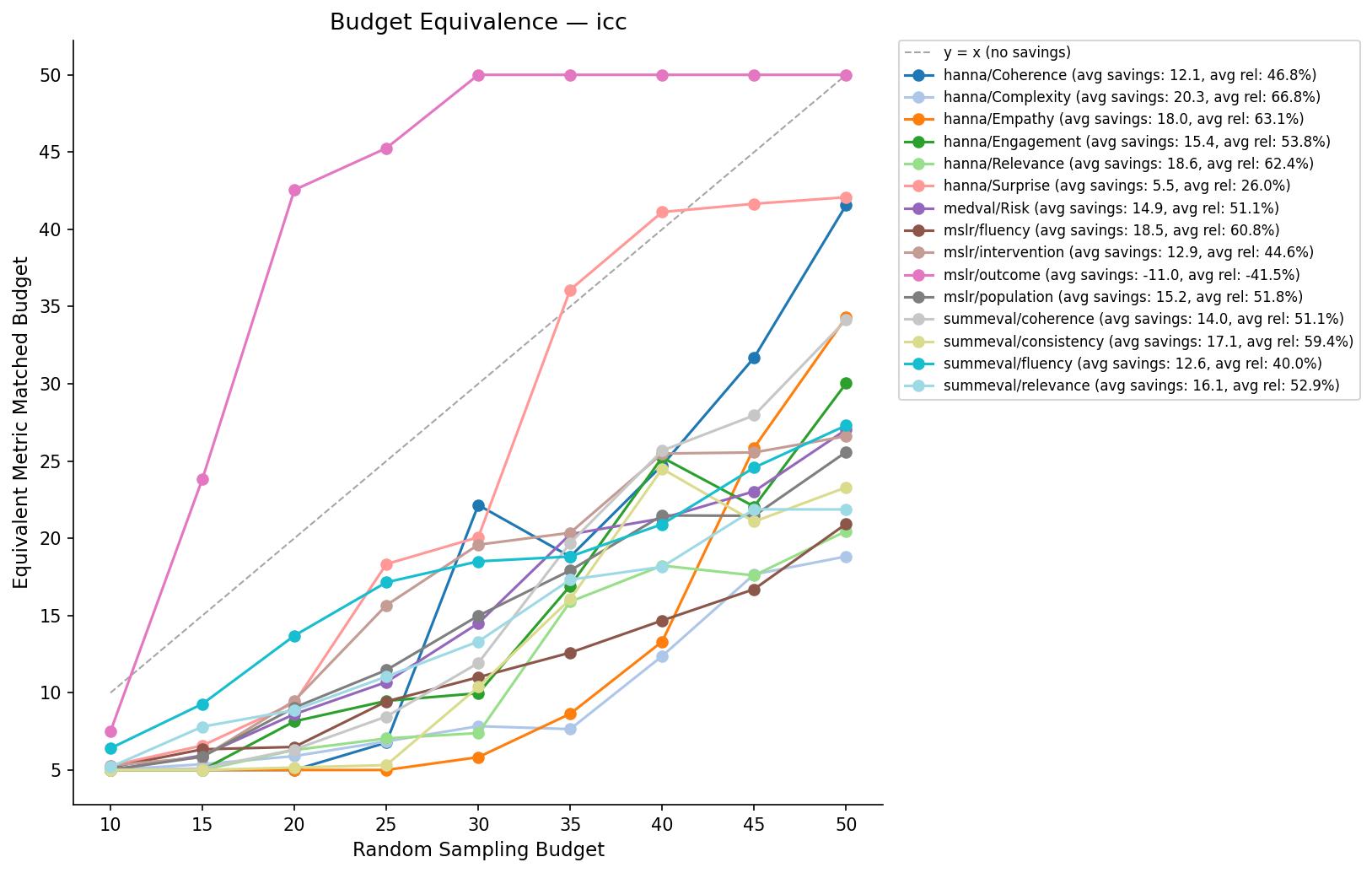}
    \end{subfigure}
    \hfill
    \begin{subfigure}{0.45\textwidth}
        \centering
        \includegraphics[width=\linewidth]{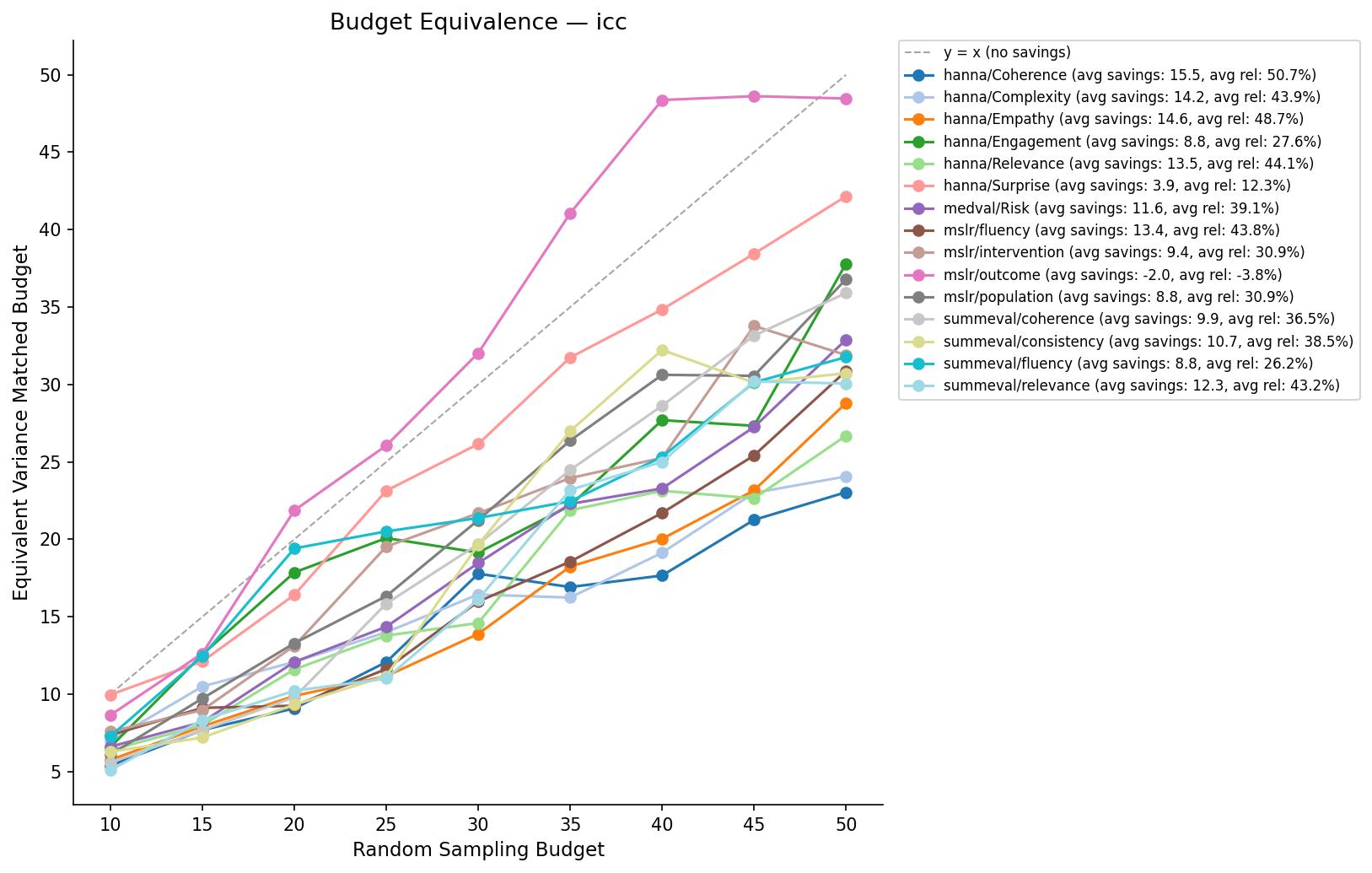}
    \end{subfigure}
        \begin{subfigure}{0.45\textwidth}
        \centering
        \includegraphics[width=\linewidth]{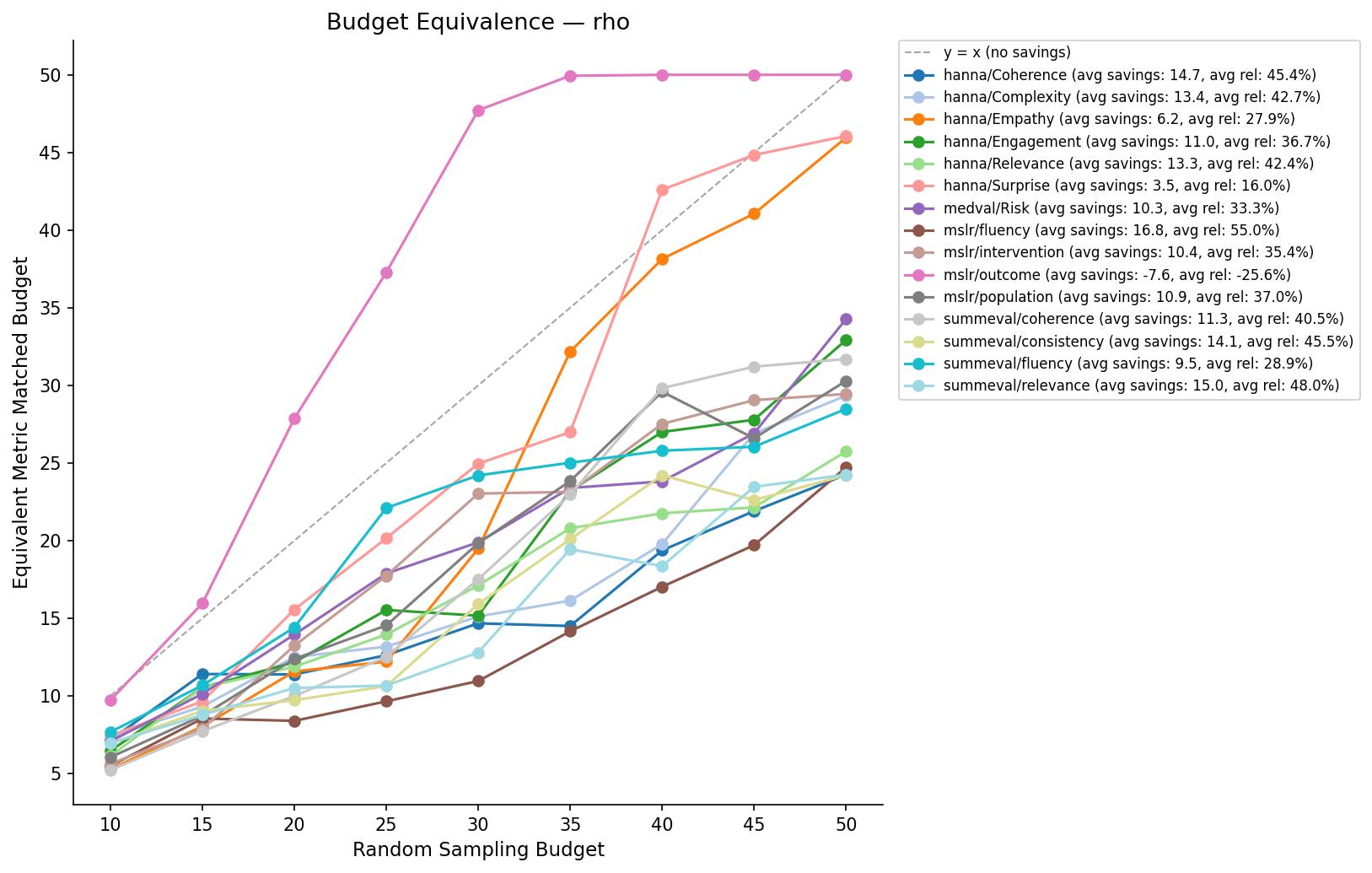}
    \end{subfigure}
    \hfill
    \begin{subfigure}{0.45\textwidth}
        \centering
        \includegraphics[width=\linewidth]{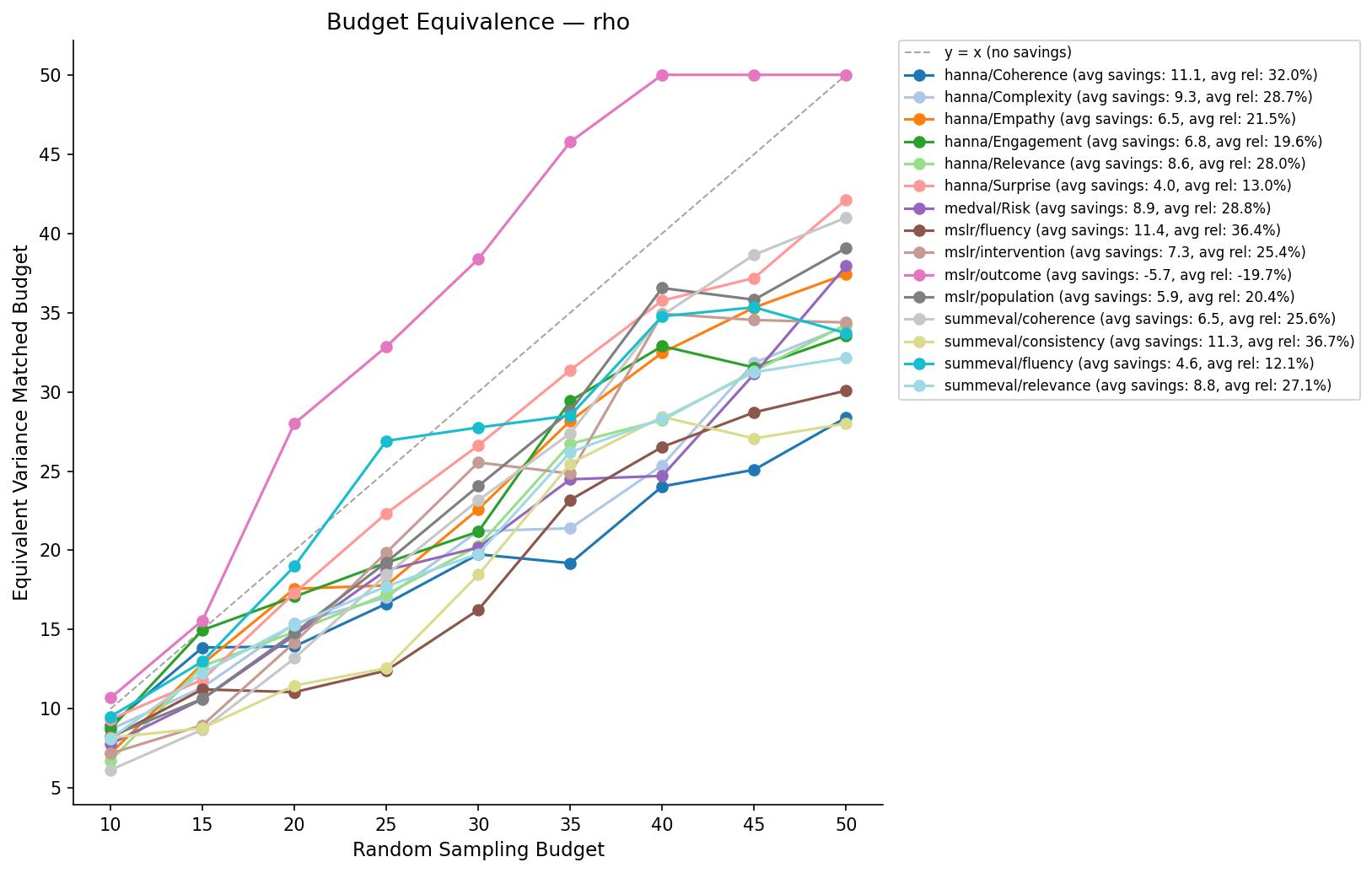}
    \end{subfigure}
        \begin{subfigure}{0.45\textwidth}
        \centering
        \includegraphics[width=\linewidth]{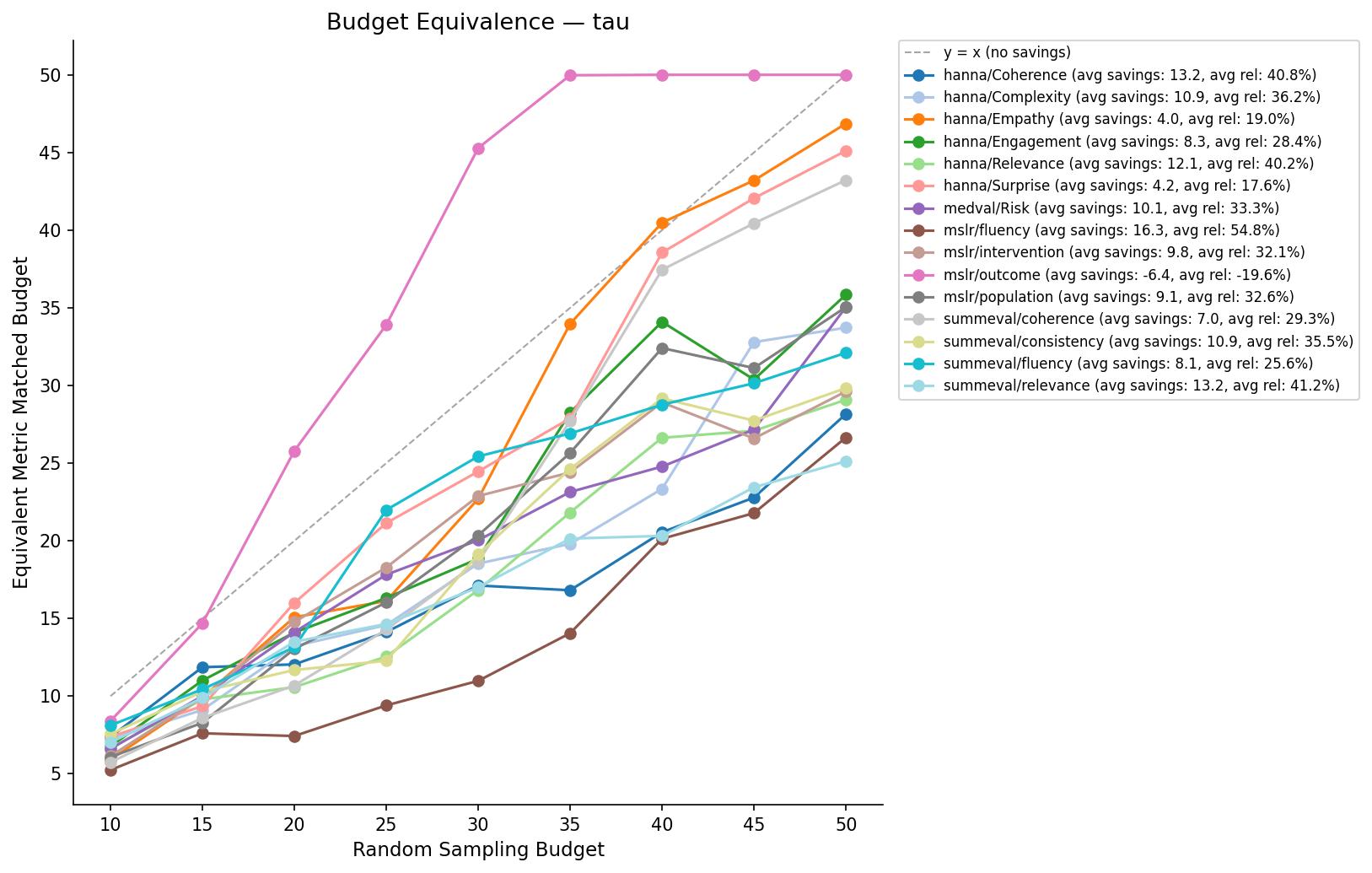}
    \end{subfigure}
    \hfill
    \begin{subfigure}{0.45\textwidth}
        \centering
        \includegraphics[width=\linewidth]{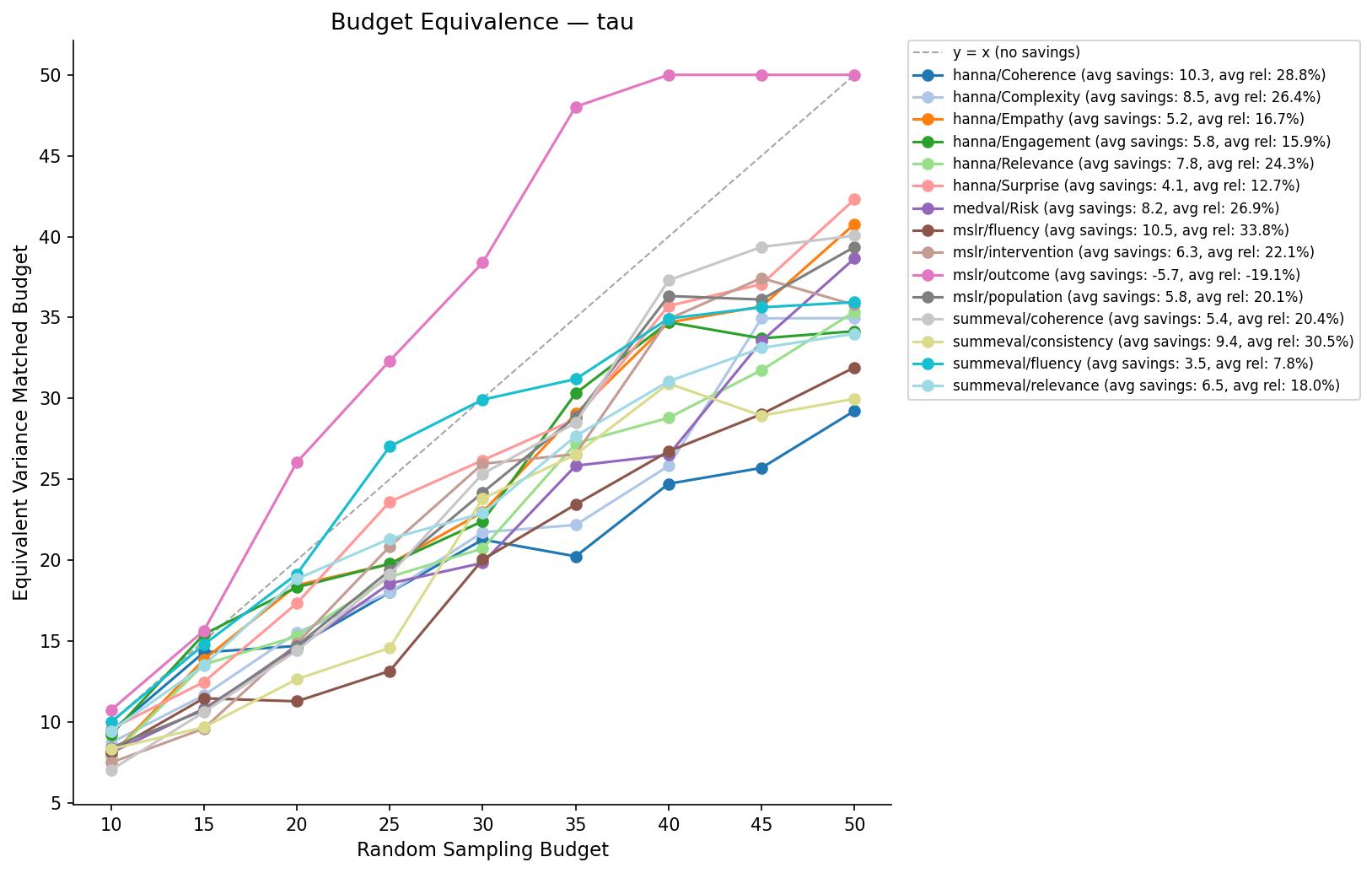}
    \end{subfigure}

    \caption{Annotation savings for both \textbf{Metric Match} as wel as associated \textbf{Variance Match} selection methods.}
    \label{fig:budget_equiv_summary}
\end{figure}

We also highlight the true reliability coefficient value per (dataset,axis) pair in Table~\ref{tab:true}, along with the estimation error (Table~\ref{tab:est_dis}) and reliability classification (Table~\ref{tab:rel_dis}) win rate between \textbf{Metric Match} and random selection at the individual dataset level.

\begin{table}[H]
\centering
\begin{tabular}{llrrrrrr}
\hline
Dataset & Axis & Alpha & ICC & Rho & Tau & MSE \\
\hline
hanna & Coherence & -0.230 & 0.679 & 0.456 & 0.391 & 4.275 \\
hanna & Complexity & 0.181 & 0.605 & 0.332 & 0.277 & 1.923 \\
hanna & Empathy & 0.194 & 0.526 & 0.302 & 0.256 & 1.688 \\
hanna & Engagement & 0.025 & 0.671 & 0.404 & 0.344 & 2.592 \\
hanna & Relevance & 0.404 & 0.755 & 0.603 & 0.516 & 1.700 \\
hanna & Surprise & 0.128 & 0.246 & 0.197 & 0.160 & 3.103 \\
medval & Risk & 0.655 & 0.812 & 0.706 & 0.629 & 1.196 \\
mslr & fluency & 0.625 & 0.779 & 0.655 & 0.643 & 0.169 \\
mslr & intervention & 0.541 & 0.700 & 0.538 & 0.487 & 0.556 \\
mslr & outcome & -0.113 & -0.005 & 0.000 & -0.000 & 1.430 \\
mslr & population & 0.377 & 0.545 & 0.372 & 0.341 & 0.642 \\
summeval & coherence & 0.644 & 0.787 & 0.668 & 0.548 & 1.143 \\
summeval & consistency & 0.654 & 0.812 & 0.762 & 0.718 & 0.987 \\
summeval & fluency & -0.263 & 0.522 & 0.420 & 0.365 & 4.222 \\
summeval & relevance & 0.527 & 0.774 & 0.576 & 0.481 & 0.951 \\
\hline
\end{tabular}
\caption{True population values of disagreggated datasets across metrics, averaged over our model suite.}
\label{tab:true}
\end{table}

\begin{table}[H]
\centering
\begin{tabular}{llrrrrr}
\hline
Dataset & Axis & Alpha & ICC & Rho & Tau & MSE \\
\hline
hanna & Coherence & 0.00 & 0.76 & 0.94 & 0.94 & 0.00 \\
hanna & Complexity & 1.00 & 1.00 & 1.00 & 0.94 & 0.80 \\
hanna & Empathy & 1.00 & 0.90 & 0.72 & 0.66 & 0.78 \\
hanna & Engagement & 0.82 & 0.92 & 0.90 & 0.84 & 0.24 \\
hanna & Relevance & 0.70 & 0.98 & 0.96 & 0.94 & 0.48 \\
hanna & Surprise & 0.38 & 0.68 & 0.76 & 0.70 & 0.66 \\
medval & Risk & 0.98 & 1.00 & 1.00 & 0.94 & 0.84 \\
mslr & fluency & 1.00 & 1.00 & 0.98 & 1.00 & 0.96 \\
mslr & intervention & 0.90 & 1.00 & 0.94 & 0.94 & 0.90 \\
mslr & outcome & 0.10 & 0.20 & 0.18 & 0.30 & 0.14 \\
mslr & population & 0.96 & 1.00 & 0.96 & 0.94 & 1.00 \\
summeval & coherence & 1.00 & 0.96 & 0.94 & 0.84 & 0.92 \\
summeval & consistency & 0.96 & 1.00 & 0.98 & 0.98 & 0.94 \\
summeval & fluency & 0.08 & 0.98 & 0.94 & 0.94 & 0.00 \\
summeval & relevance & 0.98 & 1.00 & 0.98 & 0.94 & 0.82 \\
\hline
\end{tabular}
\caption{Disagreggated results for estimation error win rate of \textbf{Metric Match} compared to random selection.}
\label{tab:est_dis}
\end{table}

\begin{table}[H]
\centering
\begin{tabular}{llrrrrr}
\hline
Dataset & Axis & Alpha & ICC & Rho & Tau & MSE \\
\hline
hanna & Coherence & 0.00 & 0.29 & 0.83 & 0.93 & 0.00 \\
hanna & Complexity & 0.85 & 0.47 & 0.70 & 0.81 & 0.70 \\
hanna & Empathy & 0.49 & 0.19 & 0.50 & 0.51 & 0.56 \\
hanna & Engagement & 0.56 & 0.56 & 0.71 & 0.69 & 0.17 \\
hanna & Relevance & 0.59 & 0.63 & 0.89 & 0.88 & 0.10 \\
hanna & Surprise & 0.49 & 0.43 & 0.70 & 0.76 & 0.64 \\
medval & Risk & 0.67 & 0.91 & 0.65 & 0.76 & 0.40 \\
mslr & fluency & 0.77 & 0.67 & 0.87 & 0.84 & 1.00 \\
mslr & intervention & 0.72 & 0.64 & 0.80 & 0.82 & 0.66 \\
mslr & outcome & 0.36 & 0.06 & 0.38 & 0.44 & 0.00 \\
mslr & population & 0.85 & 0.84 & 0.86 & 0.94 & 0.73 \\
summeval & coherence & 0.82 & 0.74 & 0.73 & 0.75 & 0.42 \\
summeval & consistency & 0.38 & 0.61 & 0.63 & 0.67 & 0.55 \\
summeval & fluency & --- & 0.26 & 0.80 & 0.83 & 0.00 \\
summeval & relevance & 0.73 & 0.60 & 0.64 & 0.71 & 0.68 \\
\hline
\end{tabular}
\caption{Disagreggated results for reliability classification win rate of \textbf{Metric Match} compared to random selection.}
\label{tab:rel_dis}
\end{table}

\section{Target and Ensemble Ablations}
\label{subsec:small_models}
\subsection{Small models}
Here we present the same analysis but with a smaller, less powerful suite of models to show generalizability. Models include GPT-4o-mini\cite{gpt}, Meta-Llama-3.1-8B-Instruct\cite{llama}, Google-gemma-3-1b-it\cite{gemma}, and Qwen2.5-7B-Instruct\cite{qwen}.

\begin{table}[H]
\centering
\begin{tabular}{lrrrrrrrrrr}
\hline
& \multicolumn{5}{c}{Estimation Error} & \multicolumn{5}{c}{Thresholding} \\
\cmidrule(lr){2-6} \cmidrule(lr){7-11}
Budget & Alpha & ICC & Rho & Tau & MSE & Alpha & ICC & Rho & Tau & MSE \\
\hline
5 & 0.92 & 0.90 & 0.91 & 0.95 & 0.85 & 0.74 & 0.48 & 0.77 & 0.81 & 0.68 \\
10 & 0.80 & 0.92 & 0.81 & 0.84 & 0.72 & 0.92 & 0.59 & 0.89 & 0.91 & 0.79 \\
15 & 0.80 & 0.88 & 0.82 & 0.84 & 0.68 & 0.84 & 0.62 & 0.94 & 0.96 & 0.79 \\
20 & 0.78 & 0.85 & 0.83 & 0.81 & 0.62 & 0.88 & 0.61 & 0.83 & 0.96 & 0.66 \\
25 & 0.78 & 0.83 & 0.88 & 0.86 & 0.62 & 1.00 & 0.58 & 0.95 & 1.00 & 0.80 \\
30 & 0.78 & 0.85 & 0.86 & 0.84 & 0.53 & 0.92 & 0.85 & 0.92 & 1.00 & 0.73 \\
35 & 0.82 & 0.87 & 0.84 & 0.79 & 0.55 & 1.00 & 0.77 & 0.88 & 1.00 & 0.81 \\
40 & 0.73 & 0.92 & 0.84 & 0.79 & 0.50 & 1.00 & 0.76 & 0.94 & 1.00 & 0.76 \\
45 & 0.70 & 0.90 & 0.78 & 0.79 & 0.47 & 1.00 & 0.84 & 0.67 & 1.00 & 0.73 \\
50 & 0.73 & 0.85 & 0.74 & 0.74 & 0.45 & 1.00 & 0.75 & 1.00 & --- & 0.81 \\
\hline
Average & 0.78 & 0.88 & 0.83 & 0.83 & 0.60 & 0.93 & 0.69 & 0.88 & 0.96 & 0.76 \\
\hline
\end{tabular}
\caption{Estimation Error and Thresholding by Budget w/ small ensemble}
\label{tab:combined_budget}
\end{table}

\begin{figure}[H]
    \centering

    \begin{subfigure}[b]{0.45\textwidth}
        \centering
        \includegraphics[width=\textwidth]{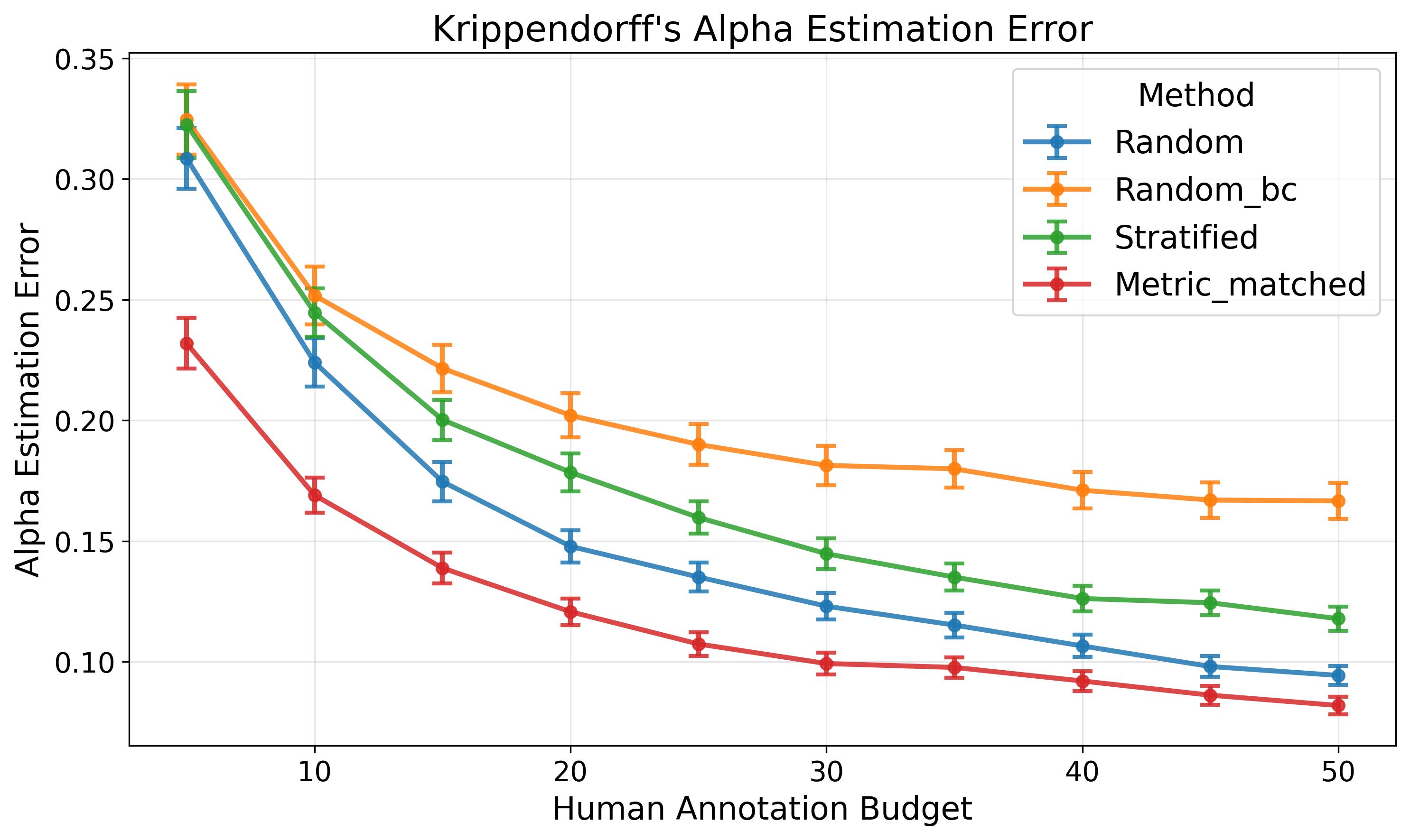}
    \end{subfigure}
    \hfill
    \begin{subfigure}[b]{0.45\textwidth}
        \centering
        \includegraphics[width=\textwidth]{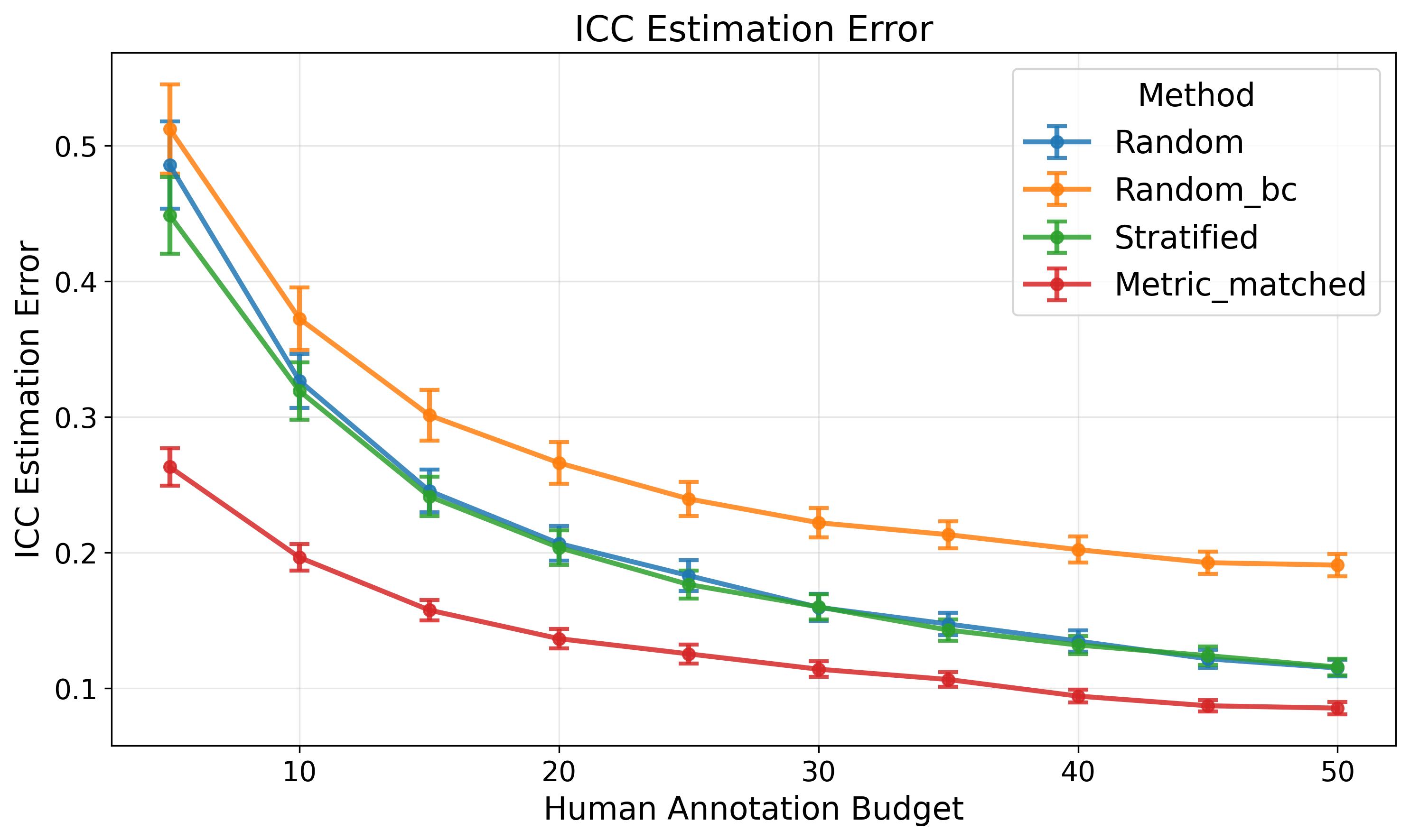}
    \end{subfigure}

    \vspace{0.5cm}

    \begin{subfigure}[b]{0.45\textwidth}
        \centering
        \includegraphics[width=\textwidth]{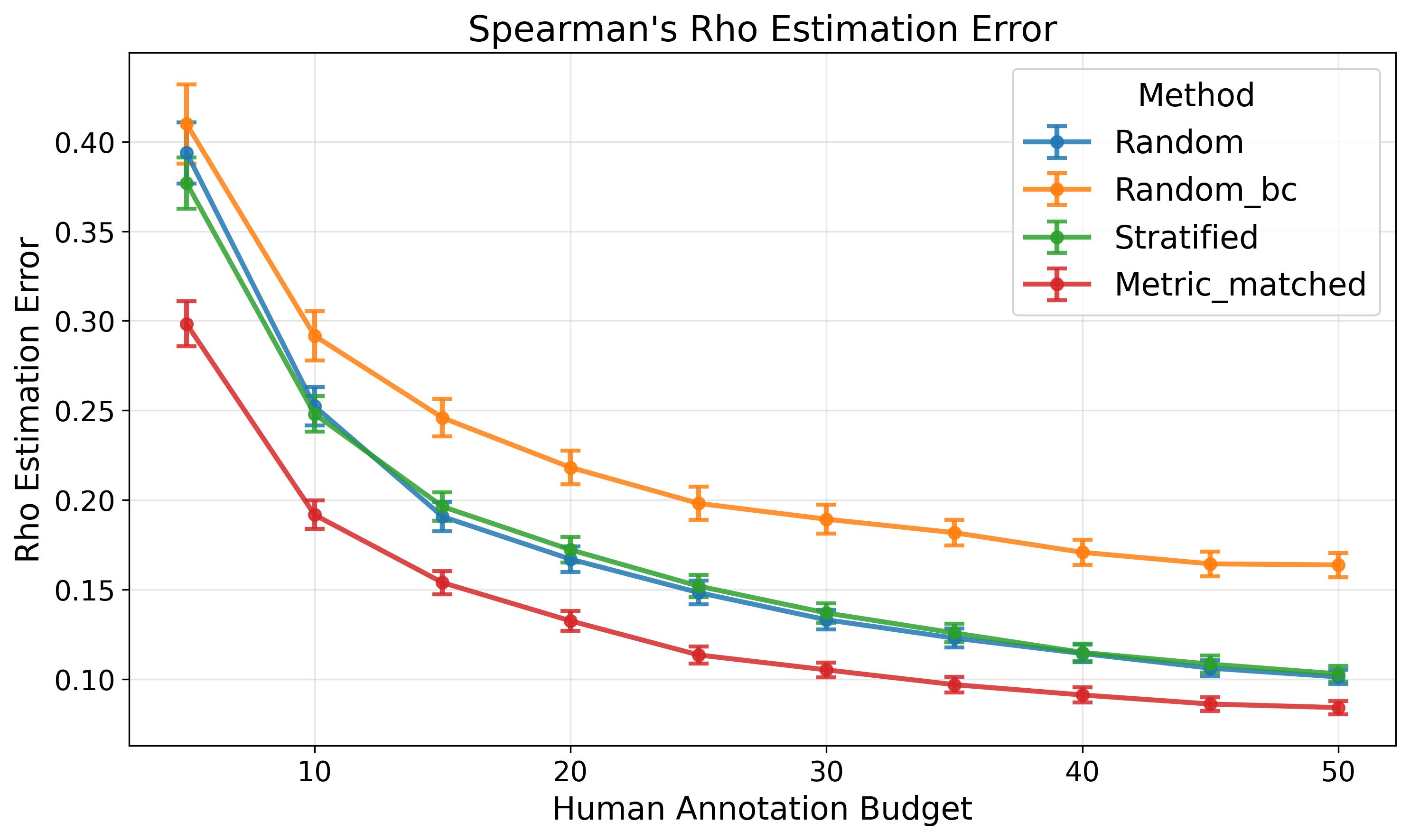}
    \end{subfigure}
    \hfill
    \begin{subfigure}[b]{0.45\textwidth}
        \centering
        \includegraphics[width=\textwidth]{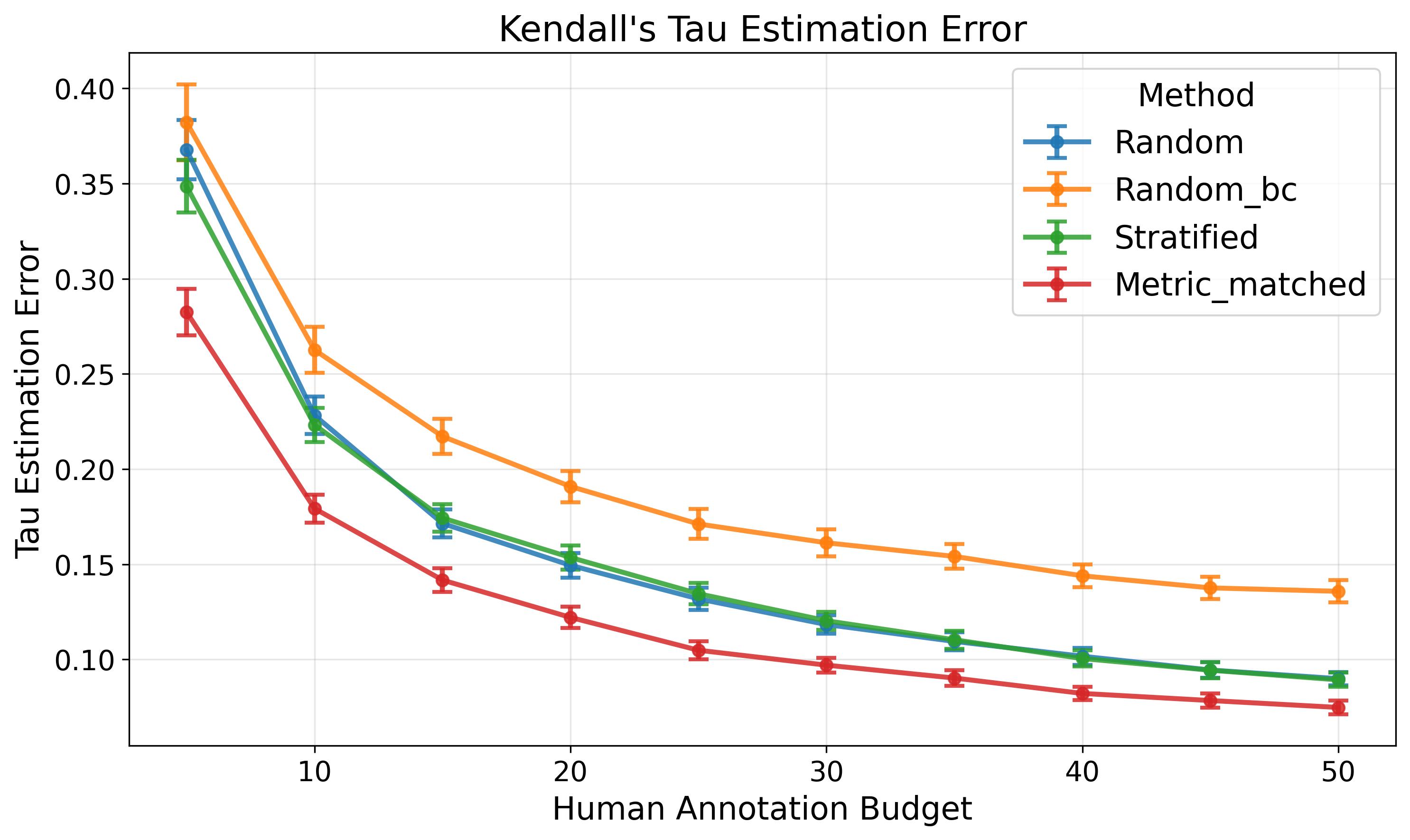}
    \end{subfigure}
        \vspace{0.5cm}

    \caption{\textbf{Small models show similar performance gains across metrics and budgets.} We report results from a suite of small models to highlight that this approach generalizes across model scale.}
    \label{fig:four_subfigures}
\end{figure}

\subsection{Single Model Ensemble Ablations}
We provide additional ablations for changing the ensemble itself that provides synthetic labels for selection of points to annotate. We see that the largest gains over random are present when an ensemble of models are used as opposed to individual models, which are more prone to reliance on the correlation of any individual model with human feedback for a given dataset. Ensembling in this setting helps improve robustness. Further, we identify that the target models must be roughly aligned capability-wise with their ensemble, and having an ensemble that is miscalibrated to the target results in worse performance.

\begin{table}[h]
\centering
\caption{\textbf{Percent Relative Improvement over Random by Method and Metric}. These results  highlight that single model ensembles don't generalize as well as multi-model ensembles.}
\begin{tabular}{lrrrrr}
\toprule
\textbf{Metric} & \makecell{\textbf{Strong Ens.+} \\ \textbf{Strong Target}} & \textbf{Claude-Only} & \textbf{GPT-5-Only} & \makecell{\textbf{Weak Ens.+} \\ \textbf{Strong Target}} & \makecell{\textbf{Weak Ens.+} \\ \textbf{Weak Target}} \\
\midrule
ICC    & 26.65\% & 15.71\% & 23.56\% & $-$34.49\% & 33.09\% \\
Alpha  &  9.05\% &  -2.88\% &  0.80\% & $-$46.87\% & 21.56\% \\
Rho    & 20.95\% & 9.61\% & 13.57\% & $-$19.26\% & 21.67\% \\
Tau    & 17.93\% & 5.74\% & 11.12\% & $-$18.36\% & 20.11\% \\
\bottomrule
\end{tabular}
\label{tab:relative_improvement}
\end{table}

\clearpage

\end{document}